\documentclass{article}
\usepackage{subcaption}

\usepackage{arxiv}
\usepackage[utf8]{inputenc} 
\usepackage[T1]{fontenc}    
\usepackage[hidelinks]{hyperref}       
\usepackage{url}            
\usepackage{booktabs}       
\usepackage{amsfonts}       
\usepackage{amsmath}        
\usepackage{amssymb}        
\usepackage{nicefrac}       
\usepackage{microtype}      
\usepackage{graphicx}
\usepackage[numbers]{natbib}

\usepackage{doi}
\usepackage{tabularx}
\usepackage{newtxtext}

\usepackage{xcolor}
\usepackage{colortbl}
\usepackage{caption}
\usepackage{algorithm}
\usepackage{algpseudocode}
\usepackage{float}
\usepackage[section]{placeins}

\newcommand{\best}[1]{\colorbox{yellow!50}{\textbf{#1}}}

\title{ForamDeepSlice: A High-Accuracy Deep Learning Framework for Foraminifera Species Classification from 2D Micro-CT Slices}

\author{
\href{https://orcid.org/0009-0008-9938-4783}{\includegraphics[scale=0.06]{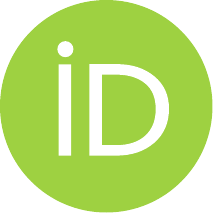}\hspace{1mm}Abdelghafour Halimi} \\
Visualization Core Lab \\
Thuwal, 23955, Saudi Arabia \\
\texttt{abdelghafour.halimi@kaust.edu.sa} \\
\And
\href{https://orcid.org/0009-0000-5047-0527}{\includegraphics[scale=0.06]{orcid.pdf}\hspace{1mm}Ali Alibrahim} \\
Physical Sciences and Engineering Division \\
Thuwal, 23955, Saudi Arabia \\
\texttt{ali.alibrahim@kaust.edu.sa} \\
\And
\href{https://orcid.org/0000-0002-6464-6607}{\includegraphics[scale=0.06]{orcid.pdf}\hspace{1mm}Didier Barradas-Bautista} \\
Visualization Core Lab \\
Thuwal, 23955, Saudi Arabia \\
\texttt{didier.barradas@kaust.edu.sa} \\
\And
\href{https://orcid.org/0000-0001-7037-1614}{\includegraphics[scale=0.06]{orcid.pdf}\hspace{1mm}Ronell Sicat} \\
Visualization Core Lab \\
Thuwal, 23955, Saudi Arabia \\
\texttt{ronell.sicat@kaust.edu.sa} \\
\And
\href{https://orcid.org/0000-0002-3193-4792}{\includegraphics[scale=0.06]{orcid.pdf}\hspace{1mm}Abdulkader M. Afifi} \\
Physical Sciences and Engineering Division \\
Thuwal, 23955, Saudi Arabia \\
\texttt{abdulkader.alafifi@kaust.edu.sa} \\
}

\renewcommand{\shorttitle}{\textit{Foraminifera Classification using Deep Learning}}

\hypersetup{
pdftitle={ForamDeepSlice},
pdfsubject={Machine learning, Microfossil},
pdfauthor={Halimi, Barradas-Bautista},
pdfkeywords={foraminifera classification, deep learning, transfer learning, 2D slices, 3D micro-CT data},
}

\begin{document}
\maketitle

\begin{abstract}
This study presents a comprehensive deep learning pipeline for the automated classification of foraminifera species using 2D micro-CT slices derived from 3D scans. We curated a scientifically rigorous dataset of 97 micro-CT scanned specimens spanning 27 species, from which we selected 12 representative species with sufficient specimen counts (at least four 3D models each) for robust classification. To ensure methodological integrity and prevent data leakage, we employed specimen-level data splitting, resulting in 109{,}617 high-quality 2D slices (44{,}103 for training, 14{,}046 for validation, and 51{,}468 for testing). We evaluated seven state-of-the-art 2D convolutional neural network (CNN) architectures using transfer learning. Our final ensemble model, \textit{ForamDeepSlice} (FDS), combining ConvNeXt-Large and EfficientNetV2-Small, achieved a test accuracy of 95.64\%, with a top-3 accuracy of 99.6\% and an area under the ROC curve (AUC) of 0.998 across all species. To facilitate practical deployment, we developed an interactive advanced dashboard that supports real-time slice classification and 3D slice matching using advanced similarity metrics, including SSIM, NCC, and the Dice coefficient. This work establishes new benchmarks for AI-assisted micropaleontological identification and provides a fully reproducible framework for foraminifera classification research, bridging the gap between deep learning and applied geosciences.
\end{abstract}


\section{Introduction}

Paleontology is undergoing a transformative data revolution, catalyzed by the widespread adoption of high-resolution, non-destructive imaging technologies such as micro-Computed Tomography (micro-CT) \cite{knutsen_accelerating_2024, edie_high-throughput_2023}. These imaging techniques have become indispensable for exploring the internal morphology of fossil specimens, offering unprecedented access to delicate anatomical structures that would otherwise be compromised by traditional mechanical preparation techniques \cite{hermanova_benefits_2020,cunningham_virtual_2014}. However, this technological leap has introduced a significant bottleneck in post-processing: the manual segmentation of fossils from surrounding rock matrices. As imaging resolutions increase, the volumetric datasets grow exponentially, making segmentation, especially for low-contrast specimens like calcareous fossils in carbonate-rich matrices, the most labor-intensive and error-prone step in the workflow \cite{edie_high-throughput_2023}.

To address this challenge, the field has increasingly turned to deep learning, a subset of machine learning that excels in automated image analysis. Early applications demonstrated that convolutional neural networks (CNNs), particularly U-Net architectures, could segment fossils with accuracies rivaling manual annotation, but in a fraction of the time \cite{knutsen_accelerating_2024, edie_high-throughput_2023}.

Building on these foundations, semantic segmentation of microfossils has advanced through hybrid architectures combining U-Net and ResNet, enabling precise segmentation in CT images with IoU scores exceeding 94\%~\cite{hou_semantic_2021} and automated identification and segmentation of microfossils in carbonate rock matrices from MicroCT data~\cite{carvalho_automated_2020}. These methods preserve spatial fidelity and support detailed reconstructions essential for paleoenvironmental and geological analyses.

These breakthroughs in automated segmentation have laid the foundation for a broader integration of artificial intelligence (AI) into paleontological research, extending beyond segmentation to include fossil classification and reconstruction, as summarized in Table~\ref{tab:sota_applications_fullwidth}. One of the most promising directions in this domain is the classification of microfossils using transfer learning. Several studies have shown that pretrained CNN architectures, such as VGG16, ResNet50, MobileNetV2, InceptionV3, and Xception, can be fine-tuned to classify microfossils from low-resolution optical images with remarkable accuracy. For instance, classification of the genus \textit{Globotruncanita} achieved up to 96.97\% accuracy and an AUC of 0.978 \cite{ozer_exploration_2024}, while broader genus-level classification reached 99.78\% accuracy \cite{ozer_towards_2023}. These approaches are particularly valuable for laboratories with limited computational resources, offering high performance with minimal infrastructure.

\begin{table}[!htbp]
\centering
\caption{Summary of State-of-the-Art Deep Learning Applications in Fossil Analysis.}
\label{tab:sota_applications_fullwidth}
\vspace{0.5em}
\renewcommand{\arraystretch}{1.4}
\setlength{\tabcolsep}{10pt}
\begin{tabularx}{\textwidth}{@{}X X X X X@{}}
\toprule
\textbf{Reference} & \textbf{Fossil Type} & \textbf{Imaging Modality / Data Type} & \textbf{Task / Model} & \textbf{Key Contribution} \\
\midrule
Knutsen et al., 2024 \cite{knutsen_accelerating_2024} & Vertebrate limb bone & Synchrotron micro-CT & Segmentation / U-Net & Dice similarity of 0.96 with $<$2\% training data \\
Edie et al., 2023 \cite{edie_high-throughput_2023} & Calcareous shelly invertebrates & Micro-CT & Segmentation / Deep Learning & Workflow for low-contrast fossil segmentation \\
Ferreira et al., 2023 \cite{10171244} & Foraminifera microfossils & 2D Images & Synthetic Data / ViT + StyleGAN & High-quality synthetic datasets for classification and segmentation \\
Mimura et al., 2025 \cite{mimura_classifying_2025} & Radiolarians (32 classes) & Optical Microscopy & Vision Transformers / ViT + FDSL & Accuracy of 86.3\%; fractal-pretrained synthetic data matches real image performance \\
Hou et al., 2020 \cite{hou_admorph_2020} & Vertebrate microfossils & 3D Models & Dataset + Benchmark / VGG16 + LSTM-RNN + SVM & ADMorph: 3D digital microfossil morphology dataset with classification benchmarks \\
Hou et al., 2021 \cite{hou_semantic_2021} & Fish microfossils & CT Images & Semantic Segmentation / U-Net + ResNet34 & Precise segmentation of fish microfossils \\
Hou et al., 2023 \cite{hou_fossil_2023} & Fusulinid microfossils & 2D Images & Classification / Multi-view Ensemble & High agreement with experts; robust with limited data \\
Ozer et al., 2023 \cite{ozer_species-level_2023} & Globotruncana (3 species) & Optical Microscopy & Hybrid Model / CNN + LSTM / BiLSTM & Accuracy of 97.35\%; AUC of 0.968 \\
Ozer et al., 2023 \cite{ozer_towards_2023} & Globotruncana and Globotruncanita & 2D Optical Images & Transfer Learning / ResNet50, Xception, InceptionV3, VGG16, MobileNet & Genus-level classification with 99.78\% accuracy; species-level 81.19\% \\
Ozer et al., 2024 \cite{ozer_exploration_2024} & Globotruncanita (3 species) & Optical Microscopy & Transfer Learning / VGG16, ResNet50, InceptionV3, MobileNetV2, Xception & Accuracy of 96.97\%; AUC of 0.978 \\
Carvalho et al., 2020 \cite{carvalho_automated_2020} & Carbonate rock microfossils & MicroCT Images & Semantic Segmentation / U-Net + ResNet (18, 34, 50, 101) & Automatic identification of microfossils in depositional context \\
Itaki et al., 2020 \cite{itaki_automated_2020,itaki_innovative_2020} & Radiolarians (siliceous) & Transmitted Light (CCD, 2M and 5M pixels) & Automated Identification / CNN & Automated species picking ($>$90\% accuracy) and relative abundance estimation ($<$\,\textpm 3.2\% vs.\ expert) \\
Gorur et al., 2023 \cite{gorur_species-level_2023} & Foraminifera (Globotruncana) & Optical Microscopy (160$\times$160 px) & Supervised Classification / SVM, LDA, kNN, Custom CNN & Low-cost classification of 3 species: G. arca, G. linneiana, G. ventricosa \\
\bottomrule
\end{tabularx}
\end{table}

Beyond transfer learning, researchers have explored custom architectures built from scratch, combining CNNs with recurrent neural networks (LSTM, BiLSTM) to classify species within the genus \textit{Globotruncana}, achieving up to 97.35\% accuracy and an AUC of 0.968 \cite{ozer_species-level_2023}. Classical machine learning models such as SVM, LDA, and kNN have also demonstrated strong performance when paired with tailored CNNs, validating the utility of lightweight models in resource-constrained settings \cite{gorur_species-level_2023}.

Finally, synthetic data generation has emerged as a powerful tool to overcome the scarcity of labeled paleontological datasets. The ForamViT-GAN framework, which integrates Vision Transformers (SwinIR) with GANs (StyleGAN2), produces high-fidelity synthetic images of foraminifera \cite{10171244}. Few-shot segmentation using only five examples further demonstrates the model's efficiency. Complementary work using fractal pretraining (FDSL) with Vision Transformers has enabled multiclass classification of 32 radiolarian species with 86.3\% accuracy, showing that mathematically generated data can rival real images in training efficacy \cite{mimura_classifying_2025}.

Together, these innovations mark a paradigm shift in paleontological research, where deep learning not only accelerates traditional workflows but also opens new avenues for fossil discovery, classification, and interpretation. Among the fossil groups that stand to benefit most from these advances are foraminifera, whose exceptional species richness and diagnostic internal morphology make them both a high-value target and a challenging test case for automated classification.

Foraminifera (forams) are single-celled marine organisms that secrete calcareous tests (shells) exhibiting remarkable morphological diversity. Close to 50,000 species have been described to date, of which approximately 9,000 are extant, making them one of the most species-rich groups of shelled microorganisms in the geological record~\cite{hayward_bruce_bw_world_2025}. Their correct identification is of paramount importance to the geosciences: foraminifera serve as the primary biostratigraphic tools for determining the age of sedimentary rocks and for reconstructing paleoenvironments, including sea-level fluctuations, ocean temperatures, and depositional settings. Beyond fundamental research, accurate foraminifera identification underpins key applied disciplines: in \textit{hydrocarbon exploration}, biostratigraphic dating of well cuttings guides drilling decisions and reservoir characterization~\cite{jones_foraminifera_2014,boudagher-fadel_evolution_2018}, while in \textit{climate modeling}, foraminiferal proxy records (e.g., $\delta^{18}$O, Mg/Ca ratios) provide essential constraints on past ocean temperatures and ice-volume changes~\cite{zachos_trends_2001}. Automating species identification, therefore, has direct economic and societal relevance.

Traditionally, foraminifera embedded in lithified rocks are studied under a transmitted-light optical microscope in randomly oriented 2D thin sections (typically 30~$\mu$m thick), a petrographic technique rooted in the pioneering work of Henry Clifton Sorby on carbonate microstructure~\cite{sellwood_structure_1993} and still central to foraminiferal biostratigraphy today~\cite{ueno_carboniferous_2022}. This approach is necessary because foraminifera cannot be physically separated from the cemented rock matrix. When foraminifera can be extracted, by disaggregating unconsolidated sediments and hand-picking individual specimens under a binocular microscope, only external morphology is observed; their diagnostic internal structures, including chamber arrangement, wall microstructure, and canal systems, remain hidden~\cite{jones_foraminifera_2014,boudagher-fadel_evolution_2018}. This fundamental observational limitation has plagued micropaleontology for over a century, forcing taxonomists to infer three-dimensional architecture from sparse, randomly oriented cross-sections.

The full 3D internal morphology of foraminifera is revealed only by micro-CT scanning, a technology that produces high-resolution volumetric reconstructions of individual specimens~\cite{cunningham_virtual_2014,hermanova_benefits_2020}. However, micro-CT acquisition is inherently slow and expensive, making it impractical for the routine identification of foraminifera across the thousands of rock or sediment samples typically processed in biostratigraphic studies~\cite{edie_high-throughput_2023}. To bridge this gap, we developed a two-pronged approach: (i)~we used micro-CT to scan individual foraminifera in 3D, building a comprehensive reference library that fully characterizes the internal morphology of existing and new species; and (ii)~we trained \textit{ForamDeepSlice} (FDS), a deep learning framework, on 2D slices extracted from these 3D volumes so that it can identify foraminifera in standard 2D thin-section images, the format routinely produced in micropaleontological practice. Although this study is centered on foraminifera micro-CT data, the overall workflow is designed to be transferable to other microfossil groups and can serve as a general framework for evaluating and optimizing deep learning pipelines for automated fossil classification. Specifically, we systematically evaluate seven CNN architectures, introduce a targeted ensemble strategy (ForamDeepSlice), and present an interactive dashboard designed for practical deployment by domain scientists.

\section{Methods}

\subsection{Dataset Creation and Curation}
\begin{figure}[!htbp]
\centering
\includegraphics[width=\linewidth]{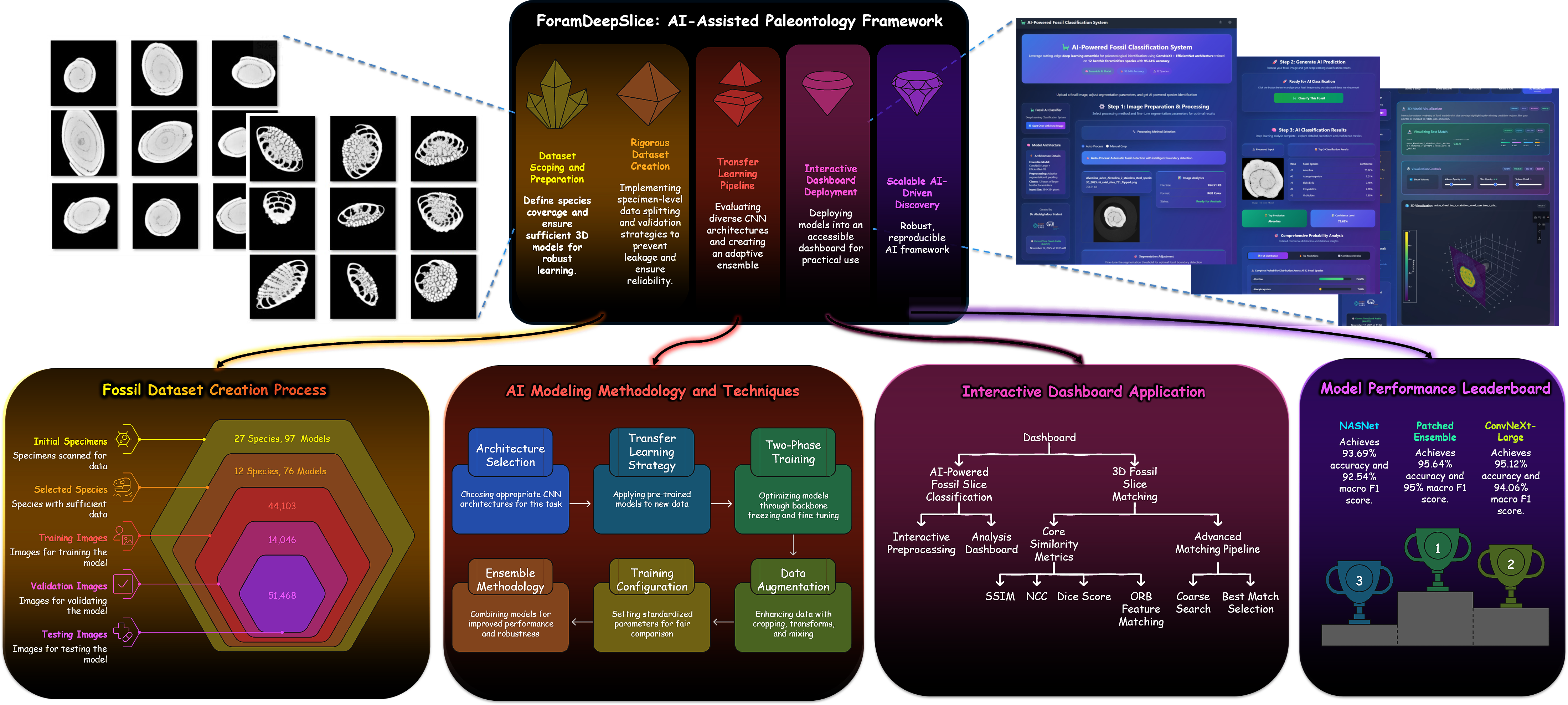}
\caption{ \textbf{Deep learning pipeline for automated foraminifera classification.}
We curated 97 micro-CT scanned specimens across 27 species, selecting 12 for robust training. Using specimen-level data splitting, 109{,}617 2D slices were processed through seven CNN architectures. Our ForamDeepSlice (FDS) model achieved 95.64\% accuracy and AUC = 0.998. An interactive dashboard enables real-time slice classification and 3D matching, setting new benchmarks for AI-assisted micropaleontology.
}
\label{fig:overview}
\end{figure}

\begin{figure}[!htbp]
\centering
\includegraphics[width=0.9\linewidth]{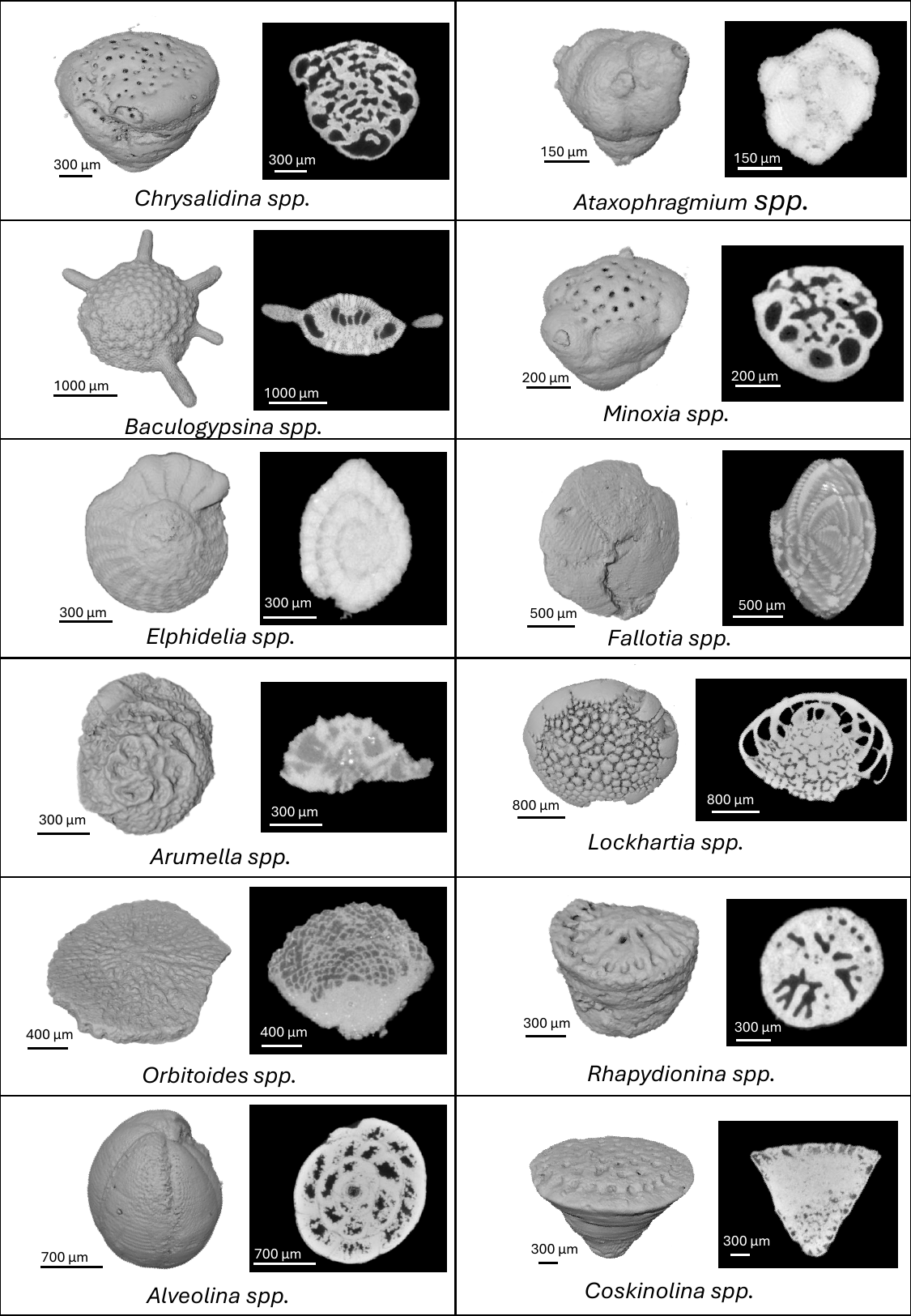}
\caption{\textbf{3D volume rendering and sample 2D slice from micro-CT data for each species used in the study.}
    We used Avizo software for the visualization and conversion of micro-CT data into NIfTI files, which are used for deep learning workflows.}
\label{fig:models}
\end{figure}

We constructed a scientifically rigorous dataset from 97 micro-CT scanned fossil specimens representing 27 foraminifera species. Twelve species with at least four 3D models each were selected to ensure robust training (we use the term 3D model to refer to individual 3D micro-CT image stacks). We provide a general overview of this AI-based workflow for the reader in Figure~\ref{fig:overview}. The selected species are summarized in Figure~\ref{fig:models} and Table~\ref{tab:species_summary}. To prevent data leakage~\cite{yagis_effect_2021}, we implemented a strict specimen-level splitting protocol, assigning all slices from a single 3D model exclusively to one of the training, validation, or test sets. This approach avoids the train-test contamination common in slice-level splits, which can overestimate accuracy by 29--55\% depending on dataset size.

Slice extraction employed Otsu's~\cite{otsu_threshold_1979} thresholding to remove low-content slices, followed by advanced segmentation to isolate fossil silhouettes. The final dataset comprises 109,617 high-quality 224$\times$224 RGB images: 44,103 for training, 14,046 for validation, and 51,468 for testing (Table~\ref{tab:species_summary}). Dataset balance was optimized using brute-force coefficient of variation minimization (Table~\ref{tab:dataset_balance}).

\begin{table}[!htbp]
\centering
\caption{Summary of species used in the study, including the number of 3D models and slice counts for training, validation, and testing. Note that we use the term 3D model to refer to a 3D micro-CT image stack.}
\label{tab:species_summary}
\vspace{0.5em}
\setlength{\tabcolsep}{12pt}
\begin{tabular}{@{}lccccc@{}}
\toprule
\textbf{Species} & \textbf{3D Models} & \textbf{Training} & \textbf{Validation} & \textbf{Test} & \textbf{Total Slices} \\
\midrule
Chrysalidina     & 16 & 3,782 & 1,079 & 5,111 & \textbf{9,972} \\
Ataxophragmium   & 7  & 3,943 & 993   & 4,898 & \textbf{9,834} \\
Baculogypsina    & 7  & 2,243 & 997   & 2,980 & \textbf{6,220} \\
Minoxia          & 6  & 4,156 & 1,034 & 3,557 & \textbf{8,747} \\
Elphidiella      & 6  & 4,048 & 1,012 & 3,385 & \textbf{8,445} \\
Fallotia         & 6  & 4,102 & 987   & 5,009 & \textbf{10,098} \\
Arumella         & 5  & 3,597 & 1,143 & 2,163 & \textbf{6,903} \\
Lockhartia       & 5  & 3,899 & 1,234 & 5,359 & \textbf{10,492} \\
Orbitoides       & 5  & 3,705 & 1,074 & 5,015 & \textbf{9,794} \\
Rhapydionina     & 5  & 3,743 & 1,186 & 3,802 & \textbf{8,731} \\
Alveolina        & 4  & 3,535 & 1,724 & 5,129 & \textbf{10,388} \\
Coskinolina      & 4  & 3,350 & 1,583 & 5,060 & \textbf{9,993} \\
\midrule
\textbf{TOTAL}   & \textbf{76} & \textbf{44,103} & \textbf{14,046} & \textbf{51,468} & \textbf{109,617} \\
\bottomrule
\end{tabular}
\end{table}

\begin{table}[!htbp]
\centering
\caption{Dataset balance metrics across training, validation, and test splits. The coefficient of variation (CV) indicates the relative variability in class distribution.}
\label{tab:dataset_balance}
\vspace{0.5em}
\setlength{\tabcolsep}{12pt}
\begin{tabular}{@{}lccc@{}}
\toprule
\textbf{Split} & \textbf{Total Images} & \textbf{CV (\%)} & \textbf{Balance Quality} \\
\midrule
Training   & 44,103 & 13.91 & Excellent \\
Validation & 14,046 & 20.58 & Good \\
Test       & 51,468 & 24.72 & Acceptable \\
\bottomrule
\end{tabular}
\end{table}

\subsubsection{Rationale for a 2D Slice-Based Approach}

Although foraminifera possess inherently three-dimensional morphologies, we adopted a 2D slice-based classification strategy for several complementary reasons. First, \textit{transfer learning availability}: the most successful deep learning architectures for image classification (e.g., ConvNeXt, EfficientNetV2, NASNet) are pre-trained on large-scale 2D image datasets such as ImageNet~\cite{deng_imagenet_2009}. No equivalently mature pre-trained 3D convolutional models exist for this domain, and training volumetric networks (e.g., 3D-ResNet, V-Net) from scratch would require substantially larger datasets than the 76 specimens (97 total scans) available here. Second, \textit{computational efficiency}: 3D CNNs demand significantly higher GPU memory and computation than their 2D counterparts; a single micro-CT volume at full resolution can exceed available VRAM, whereas individual 2D slices are efficiently processed in standard batch sizes. Third, \textit{data amplification}: slicing each 3D volume along the axial plane yields hundreds of informative cross-sections per specimen, expanding the effective dataset from 76 specimens to 109,617 labeled images (44,103 for training), a scale that enables robust fine-tuning of deep architectures. Fourth, \textit{morphological precedent}: the diagnostic features used in traditional foraminifera taxonomy (chamber arrangement, wall structure, ornamentation) are historically assessed from 2D thin-section images, and our extracted slices closely mirror this established practice.

We acknowledge that 2D slices extracted from the same 3D specimen are not statistically independent, as adjacent slices share structural continuity. This is precisely why we enforced \textit{specimen-level} data splitting rather than slice-level splitting, ensuring that all slices from a given specimen reside in exactly one partition (training, validation, or test). This protocol prevents data leakage and yields conservative, unbiased estimates of generalization performance. Nevertheless, intra-specimen slice correlations remain a limitation for deriving per-slice uncertainty estimates and should be considered when interpreting confidence scores for individual predictions.

\subsection{Data Preprocessing and Augmentation}
To enhance generalization, we applied intelligent bounding-box cropping and a suite of augmentations:
\begin{itemize}
  \item \textbf{Geometric}: rotations ($\pm$45\textdegree), scaling (0.8--1.2$\times$), horizontal flips. Crucially, the rotation augmentations simulate the random orientations encountered in traditional petrographic thin sections, where fossils are cut at arbitrary angles through the rock matrix. Because micro-CT slices are inherently axis-aligned, these rotations bridge the domain gap between digital CT data and the randomly oriented cross-sections that geologists routinely observe under a transmitted-light microscope, thereby improving the practical applicability of the trained models to real-world thin-section analysis.
  \item \textbf{Color}: brightness/contrast adjustments, jittering.
  \item \textbf{Regularization}: CutMix~\cite{yun_cutmix_2019} and Mixup~\cite{zhang_mixup_2017} strategies.
\end{itemize}

\subsection{Model Architectures and Transfer Learning Strategy}
We evaluated seven state-of-the-art 2D CNN architectures: ConvNeXt~\cite{liu_convnet_2022} (Base and Large), EfficientNetV2~\cite{tan_efficientnetv2_2021} (Small and Large), NASNet~\cite{zoph_learning_2018}, MobileNet~\cite{howard_searching_2019}, and ResNet101V2~\cite{he_deep_2016}. Transfer learning was applied using ImageNet~\cite{deng_imagenet_2009}-pretrained weights. Training followed a two-phase strategy:

\begin{itemize}
  \item \textbf{Phase 1: Backbone Freezing}, 10 epochs with frozen pretrained weights; only classifier head trained. Learning rate: 1e-4 with cosine scheduling.
  \item \textbf{Phase 2: Fine-Tuning}, 20 epochs with full network unfrozen; learning rate: 5e-5 with warm restarts. Mixed precision (fp16) training was used for efficiency. Early stopping was applied with patience=5 based on validation loss.
\end{itemize}

All models used standardized configurations: input resolution (224$\times$224 or 384$\times$384), pixel intensity normalization to $[0, 1]$, standardization using ImageNet mean and standard deviation, batch size 32, AdamW optimizer~\cite{loshchilov_decoupled_2017}, and categorical crossentropy loss with label smoothing (0.1)~\cite{szegedy_rethinking_2016}. Training was conducted on NVIDIA GPUs with fixed seeds and deterministic operations for reproducibility.

\subsection{PatchEnsemble: Confidence-Gated Model Switching}
Our final model employs a \textit{PatchEnsemble} strategy, a conditional confidence-gated model-switching mechanism that differs fundamentally from conventional ensemble methods such as majority voting, model averaging, stacking, or bagging~\cite{zhou2025ensemble}. Whereas standard ensemble approaches aggregate predictions from all constituent models for every input sample, PatchEnsemble selectively delegates predictions to a secondary ``patch'' model only when the primary model's prediction falls within a pre-identified set of difficult classes \textit{and} the patch model exhibits higher confidence. This targeted approach avoids diluting prediction quality, which can occur when weaker models contribute noise to the ensemble output for classes where the primary model already excels.

Formally, let $\mathcal{M}_{\text{main}}$ denote the primary model (ConvNeXt-Large) and $\mathcal{M}_{\text{patch}}$ denote the secondary model (EfficientNetV2-Small). For a given input image $\mathbf{x}$, each model produces a probability vector over $C$ classes:
\begin{equation}
\mathbf{p}_M = \mathcal{M}_{\text{main}}(\mathbf{x}) \in \mathbb{R}^C, \quad \mathbf{p}_P = \mathcal{M}_{\text{patch}}(\mathbf{x}) \in \mathbb{R}^C
\end{equation}
Let $\mathcal{W} \subset \{1, \ldots, C\}$ denote the set of ``weak'' class indices, identified \textit{a priori} from per-class F1-score analysis on the validation set (in this study, $\mathcal{W} = \{\textit{Baculogypsina},\; \textit{Orbitoides}\}$). The patch model was selected as the architecture achieving the highest recall on the weak classes: EfficientNetV2-Small attained the best recall for \textit{Baculogypsina} (0.497; Table~\ref{tab:recall_all}) among all seven evaluated models.

The PatchEnsemble output $\hat{\mathbf{p}}$ is defined as:
\begin{equation}
\hat{\mathbf{p}}(\mathbf{x}) =
\begin{cases}
\dfrac{\mathbf{p}_P}{\|\mathbf{p}_P\|_1} & \text{if } \arg\max(\mathbf{p}_M) \in \mathcal{W} \;\;\textbf{and}\;\; \max(\mathbf{p}_P) > \max(\mathbf{p}_M), \\[6pt]
\dfrac{\mathbf{p}_M}{\|\mathbf{p}_M\|_1} & \text{otherwise.}
\end{cases}
\end{equation}
The two conditions ensure that model switching occurs only when (1)~the primary model predicts a class known to be problematic, and (2)~the patch model is more confident in its own prediction. This conservative gating prevents the patch model from overriding correct predictions on well-classified species.

The procedure is summarized in Algorithm~\ref{alg:patchensemble}.

\begin{algorithm}[ht]
\caption{PatchEnsemble Inference}
\label{alg:patchensemble}
\begin{algorithmic}[1]
\Require Input image $\mathbf{x}$, primary model $\mathcal{M}_{\text{main}}$, patch model $\mathcal{M}_{\text{patch}}$, weak class set $\mathcal{W}$
\Ensure Final probability vector $\hat{\mathbf{p}}$
\State $\mathbf{p}_M \gets \mathcal{M}_{\text{main}}(\text{preprocess}_{\text{main}}(\mathbf{x}))$ \Comment{ConvNeXt-Large, 384$\times$384}
\State $\mathbf{p}_P \gets \mathcal{M}_{\text{patch}}(\text{preprocess}_{\text{patch}}(\mathbf{x}))$ \Comment{EfficientNetV2-S, 224$\times$224}
\State $\hat{c} \gets \arg\max(\mathbf{p}_M)$
\If{$\hat{c} \in \mathcal{W}$ \textbf{and} $\max(\mathbf{p}_P) > \max(\mathbf{p}_M)$}
    \State $\hat{\mathbf{p}} \gets \mathbf{p}_P \;/\; \|\mathbf{p}_P\|_1$ \Comment{Switch to patch model}
\Else
    \State $\hat{\mathbf{p}} \gets \mathbf{p}_M \;/\; \|\mathbf{p}_M\|_1$ \Comment{Retain primary model}
\EndIf
\State \Return $\hat{\mathbf{p}}$
\end{algorithmic}
\end{algorithm}

This design was motivated by the observation that conventional Top-$k$ ensemble methods (combining the two to four best-performing models via majority voting) failed to improve overall accuracy (Table~\ref{tab:f1_scores_all}). We attribute this to the propagation of correlated errors: when multiple models share similar failure modes on the same difficult species, aggregating their predictions amplifies rather than corrects misclassifications. By contrast, PatchEnsemble isolates the correction to only those samples where the primary model is both uncertain and incorrect on known difficult taxa, preserving its strong performance elsewhere.

With the classification pipeline established, we next describe the software infrastructure developed to deploy these models for practical use by domain scientists.

\subsection{Dashboard Architecture and Deployment}
We deployed the trained models via a dashboard with two core applications:
\begin{itemize}
  \item \textbf{Real-Time Slice Classification}: interactive interface with ranked predictions, confidence metrics, and preprocessing tools.
  \item \textbf{3D Slice Matching}: similarity analysis using Structural Similarity Index (SSIM)~\cite{zhou_wang_image_2004}, Normalized Cross-Correlation (NCC)~\cite{Lewis_Template_1995}, Dice coefficient~\cite{dice_measures_1945}, and Oriented FAST and Rotated BRIEF (ORB) feature matching~\cite{rublee_orb_2011}.
\end{itemize}

The dashboard bridges research and field application, enabling real-time fossil identification and correspondence analysis.

To maximize accessibility, we packaged the complete dashboard as a pre-built, containerized application that can be deployed with a single command on any Docker-enabled machine. The deployment uses Docker Compose to orchestrate a GPU-accelerated backend (hosting the FDS classification and 3D slice matching models) and a web-based frontend, requiring no local installation of Python, TensorFlow, or other dependencies. Platform-specific launcher scripts (\texttt{Run\_app.bat} for Windows; \texttt{start-production.sh} for Linux/macOS) are provided in the \texttt{7\_Final\_Advanced\_Application} directory of the repository. Once launched, the dashboard is accessible at \texttt{http://localhost:8080}. The following command deploys the dashboard:

\begin{verbatim}
docker compose -f docker-compose.production.yml up -d
\end{verbatim}

\subsection{Reproducibility and Environment}
All code and models are organized into modular Jupyter notebooks and Python scripts and are available in the corresponding \href{https://github.com/A-Halimi/3D_Fossil_Project}{GitHub} repository.
We provide the test dataset \href{https://repository.kaust.edu.sa/items/142ae836-df31-48c7-8a2e-11da19ffeaa5}{here}~\cite{halimi_test_2026}. For further details, please check the preprint~\cite{halimi_foramdeepslice_2025}.
A Docker environment is provided for reproducibility, with minimum system requirements: Python 3.8+, 16GB RAM, 8GB VRAM GPU, and 100GB storage. All experiments reported in this study were conducted on an NVIDIA Tesla V100-SXM2-32GB GPU using TensorFlow with mixed-precision (fp16) training enabled to maximize throughput. Training each individual model required approximately 30 epochs (10 frozen + 20 fine-tuning) with the AdamW optimizer, fixed random seeds (seed=42), and deterministic operations for full reproducibility.

The following Docker command launches the full environment required for dataset generation and model training:

\begin{verbatim}
docker run --gpus all --ipc=host --ulimit memlock=-1 --ulimit stack=67108864 \
  --rm -p 10000:8888 -p 8501:8501 -v ${PWD}:/workspace/mycode \
  abdelghafour1/ngc_tf_rapids_25_01_vscode_torch:2025-v3 \
  jupyter lab --ip=0.0.0.0 --allow-root
\end{verbatim}

\section{Results}

\subsection{Individual Model Benchmarking}
\begin{figure}[!htbp]
\centering
\includegraphics[width=\linewidth]{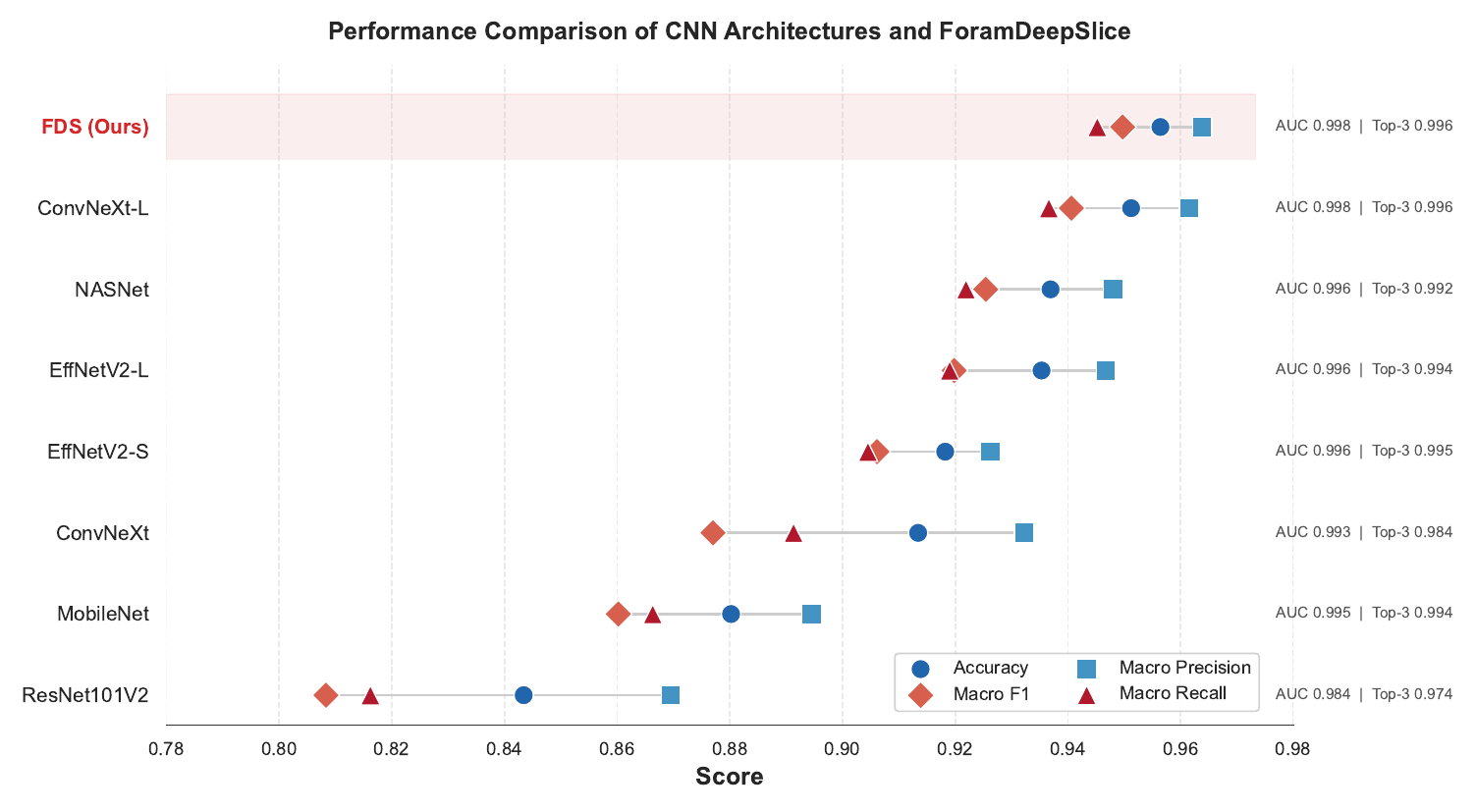}
\caption{\textbf{Performance comparison of seven state-of-the-art 2D CNN architectures and the ForamDeepSlice (FDS) ensemble.}
    Each model was evaluated on a balanced dataset of 109,617 micro-CT slices using standard classification metrics.
    Accuracy, Macro F1-score, Macro Precision, and Macro Recall are reported for each architecture, with AUC and Top-3 accuracy annotated. FDS achieved the highest overall Macro F1 (0.949) and accuracy (0.956).
    All models exhibited excellent calibration, with AUC values exceeding 0.98.}
\label{fig:model_comparison}
\end{figure}

We systematically evaluated seven state-of-the-art 2D convolutional neural network (CNN) architectures on a curated dataset comprising 109,617 micro-CT slices. To provide a clear quantitative benchmark, we report complementary metrics, namely accuracy, macro F1-score, and area under the receiver operating characteristic curve (AUC), for each model (Figure~\ref{fig:model_comparison}).

We used ResNet101V2 as a baseline and compared all other architectures against it. Classification accuracy across individual models ranged from approximately 84\% to 95\%, with ConvNeXt-Large achieving the highest top-1 accuracy among single-model configurations. Macro F1-score trends were consistent with this ranking, indicating that ConvNeXt-Large and NASNet offered the strongest precision-recall balance across species.

All models demonstrated excellent calibration, with AUC values consistently exceeding 0.98, underscoring their reliability in distinguishing between classes. Specifically, ConvNeXt-Large and EfficientNetV2-Large achieved the highest AUC of 0.998 and 0.996, respectively, while even the baseline ResNet101V2 attained an AUC of 0.984 (Table~\ref{tab:model_performance_all}). A systematic per-class analysis revealed that no single model dominated across all species: for instance, MobileNet achieved the best recall for \textit{Alveolina} (0.999), EfficientNetV2-Large led on \textit{Arumella} and \textit{Elphidiella}, and EfficientNetV2-Small attained the highest recall for \textit{Baculogypsina} (0.497; Table~\ref{tab:recall_all}), a finding that directly motivated our PatchEnsemble design (Section~\textit{ForamDeepSlice Ensemble Performance} and Algorithm~\ref{alg:patchensemble}).

\begin{table}[!htbp]
\centering
\caption{Comprehensive performance comparison across all evaluated models.}
\label{tab:model_performance_all}
\vspace{0.5em}
\renewcommand{\arraystretch}{1.4}
\setlength{\tabcolsep}{12pt}
\begin{tabular}{@{}lcccccc@{}}
\toprule
\textbf{Model} & \textbf{Accuracy} & \textbf{Precision} & \textbf{Recall} & \textbf{F1-Score} & \textbf{Top-3 Acc} & \textbf{AUC} \\
\midrule
FDS (Ours) & \best{0.956} & \best{0.964} & \best{0.945} & \best{0.950} & \best{0.996} & \best{0.998} \\
ConvNeXt-Large               &     0.951 &            0.962 &         0.937 &     0.941 & 0.996 & 0.998 \\
NASNet                  &     0.937 &            0.948 &         0.922 &     0.925 & 0.992 & 0.996 \\
EfficientNetV2-L                  &     0.935 &            0.947 &         0.919 &     0.920 & 0.994 & 0.996 \\
EfficientNetV2-S                  &     0.918 &            0.926 &         0.904 &     0.906 & 0.995 & 0.996 \\
ConvNeXt                &     0.913 &            0.932 &         0.891 &     0.877 & 0.984 & 0.993 \\
MobileNet               &     0.880 &            0.894 &         0.866 &     0.860 & 0.994 & 0.995 \\
ResNet101V2             &     0.843 &            0.870 &         0.816 &     0.808 & 0.974 & 0.984 \\
\bottomrule
\end{tabular}
\end{table}

To guide interpretation of the comprehensive performance metrics reported in Table~\ref{tab:model_performance_all}, we highlight the relevance of each metric from a paleontological and applied perspective. \textit{Macro F1-score} is the most informative single metric for this task, as it equally weights all species regardless of sample size, thereby penalizing models that perform poorly on under-represented taxa, a common scenario in micropaleontology where rare species are often the most scientifically valuable. \textit{AUC} (area under the ROC curve) captures discriminative ability across all classification thresholds and is useful for confirming model calibration; however, when all models achieve AUC $>$ 0.98, it offers limited discriminative power for model selection. \textit{Top-3 accuracy} is particularly relevant for the interactive dashboard use case, where the paleontologist reviews the top-ranked candidates rather than relying solely on the top-1 prediction; values near 99.6\% indicate that the correct species almost always appears among the top three suggestions. At the species level, the distinction between \textit{precision} and \textit{recall} carries specific paleontological meaning: low precision (high false positives) implies that the model erroneously assigns specimens to a taxon, inflating its apparent abundance, whereas low recall (high false negatives) means the model fails to detect true instances of a taxon, underestimating its stratigraphic presence. In biostratigraphic applications, false negatives for marker species may lead to incorrect age assignments, making recall the more critical metric for rare diagnostic taxa.

\subsection{Species-Level Classification Analysis}

To assess model performance at a finer granularity, we computed three key classification metrics, precision, recall, and F1-score, at the species level, treating each species as an independent class.

\begin{figure}[!htbp]
\centering
\includegraphics[width=\linewidth]{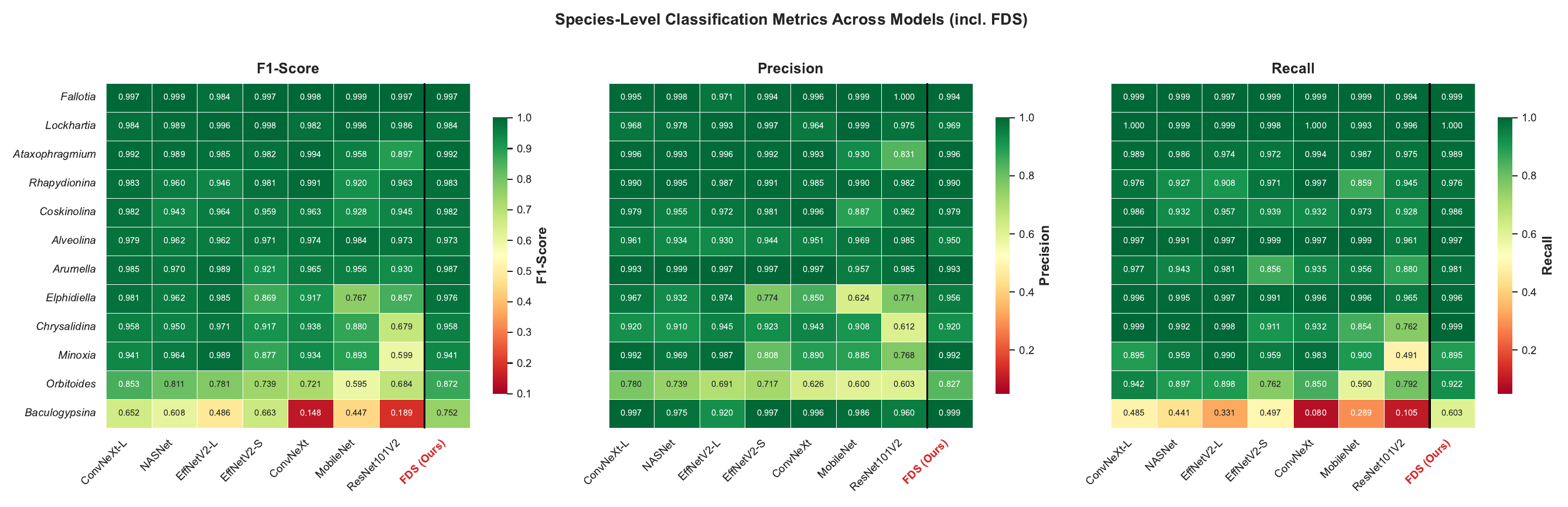}
\caption{\textbf{Species-level classification metrics across individual models and ForamDeepSlice (FDS).}
F1-score, precision, and recall were computed for each species, with species ordered from best- to worst-performing (top to bottom). The FDS column (rightmost, separated by a vertical line) shows the final ensemble performance. \textit{Baculogypsina} exhibits extremely low recall (high false negatives) while \textit{Orbitoides} suffers from low precision (high false positives), resulting in reduced F1-scores for both taxa.}
\label{fig:metrics_per_species}
\end{figure}

Most models achieved F1-scores exceeding 90\%, with \textit{Fallotia} reaching a peak performance of 99.8\% (Table~\ref{tab:f1_scores_all}). However, certain species posed greater classification challenges. Notably, \textit{Baculogypsina} and \textit{Orbitoides} exhibited the lowest F1-scores across all evaluated models (Figure~\ref{fig:metrics_per_species}).

\begin{table}[!htbp]
\centering
\caption{Comprehensive F1-scores for all species across individual models and ensemble methods. Best value per species is highlighted. FDS = ForamDeepSlice (Patched Ensemble of ConvNeXt-Large + EfficientNetV2-Small). Individual models are ordered by overall accuracy.}
\label{tab:f1_scores_all}
\vspace{0.5em}
\footnotesize
\setlength{\tabcolsep}{3pt}
\renewcommand{\arraystretch}{1.3}
\begin{tabular}{@{}l ccccccc cccc@{}}
\toprule
& \multicolumn{7}{c}{\textbf{Individual Models}} & \multicolumn{4}{c}{\textbf{Ensemble Models}} \\
\cmidrule(lr){2-8} \cmidrule(l){9-12}
\textbf{Species} & \rotatebox{70}{ConvNeXt-L} & \rotatebox{70}{NASNet} & \rotatebox{70}{EffNetV2-L} & \rotatebox{70}{EffNetV2-S} & \rotatebox{70}{ConvNeXt} & \rotatebox{70}{MobileNet} & \rotatebox{70}{ResNet101V2} & \rotatebox{70}{Top-2} & \rotatebox{70}{Top-3} & \rotatebox{70}{Top-4} & \rotatebox{70}{FDS} \\
\midrule
Alveolina        & 0.978 & 0.961 & 0.962 & 0.970 & 0.973 & \best{0.983} & 0.972 & 0.969 & 0.974 & 0.967 & 0.973 \\
Arumella         & 0.985 & 0.970 & \best{0.988} & 0.920 & 0.964 & 0.956 & 0.929 & 0.968 & 0.982 & 0.986 & 0.987 \\
Ataxophragmium   & 0.992 & 0.989 & 0.984 & 0.982 & 0.993 & 0.957 & 0.897 & 0.990 & \best{0.994} & 0.993 & 0.992 \\
Baculogypsina    & 0.652 & 0.607 & 0.486 & 0.663 & 0.148 & 0.446 & 0.189 & 0.389 & 0.366 & 0.588 & \best{0.752} \\
Chrysalidina     & 0.957 & 0.949 & 0.970 & 0.916 & 0.937 & 0.880 & 0.679 & 0.952 & \best{0.973} & \best{0.973} & 0.957 \\
Coskinolina      & \best{0.982} & 0.943 & 0.964 & 0.959 & 0.963 & 0.928 & 0.945 & 0.950 & 0.962 & 0.965 & 0.982 \\
Elphidiella      & 0.981 & 0.962 & \best{0.985} & 0.868 & 0.917 & 0.767 & 0.857 & 0.950 & 0.968 & 0.973 & 0.975 \\
Fallotia         & 0.996 & \best{0.998} & 0.983 & 0.996 & 0.997 & 0.998 & 0.996 & 0.998 & 0.992 & 0.996 & 0.996 \\
Lockhartia       & 0.983 & 0.988 & 0.996 & \best{0.997} & 0.981 & 0.995 & 0.985 & 0.984 & 0.995 & \best{0.997} & 0.983 \\
Minoxia          & 0.940 & 0.963 & \best{0.988} & 0.877 & 0.933 & 0.892 & 0.599 & 0.971 & \best{0.988} & 0.985 & 0.941 \\
Orbitoides       & 0.853 & 0.810 & 0.781 & 0.738 & 0.721 & 0.595 & 0.684 & 0.777 & 0.770 & 0.803 & \best{0.872} \\
Rhapydionina     & 0.982 & 0.959 & 0.945 & 0.981 & \best{0.990} & 0.920 & 0.962 & 0.977 & 0.972 & 0.977 & 0.982 \\
\midrule
\textbf{Macro F1}  & 0.940 & 0.925 & 0.919 & 0.906 & 0.877 & 0.860 & 0.808 & 0.906 & 0.911 & 0.934 & \best{0.949} \\
\textbf{Accuracy}  & 0.951 & 0.936 & 0.935 & 0.918 & 0.913 & 0.880 & 0.843 & 0.929 & 0.934 & 0.945 & \best{0.956} \\
\bottomrule
\end{tabular}
\end{table}

\begin{figure}[!htbp]
\centering
\begin{minipage}[t]{0.48\textwidth}
    \centering
    \includegraphics[width=\linewidth]{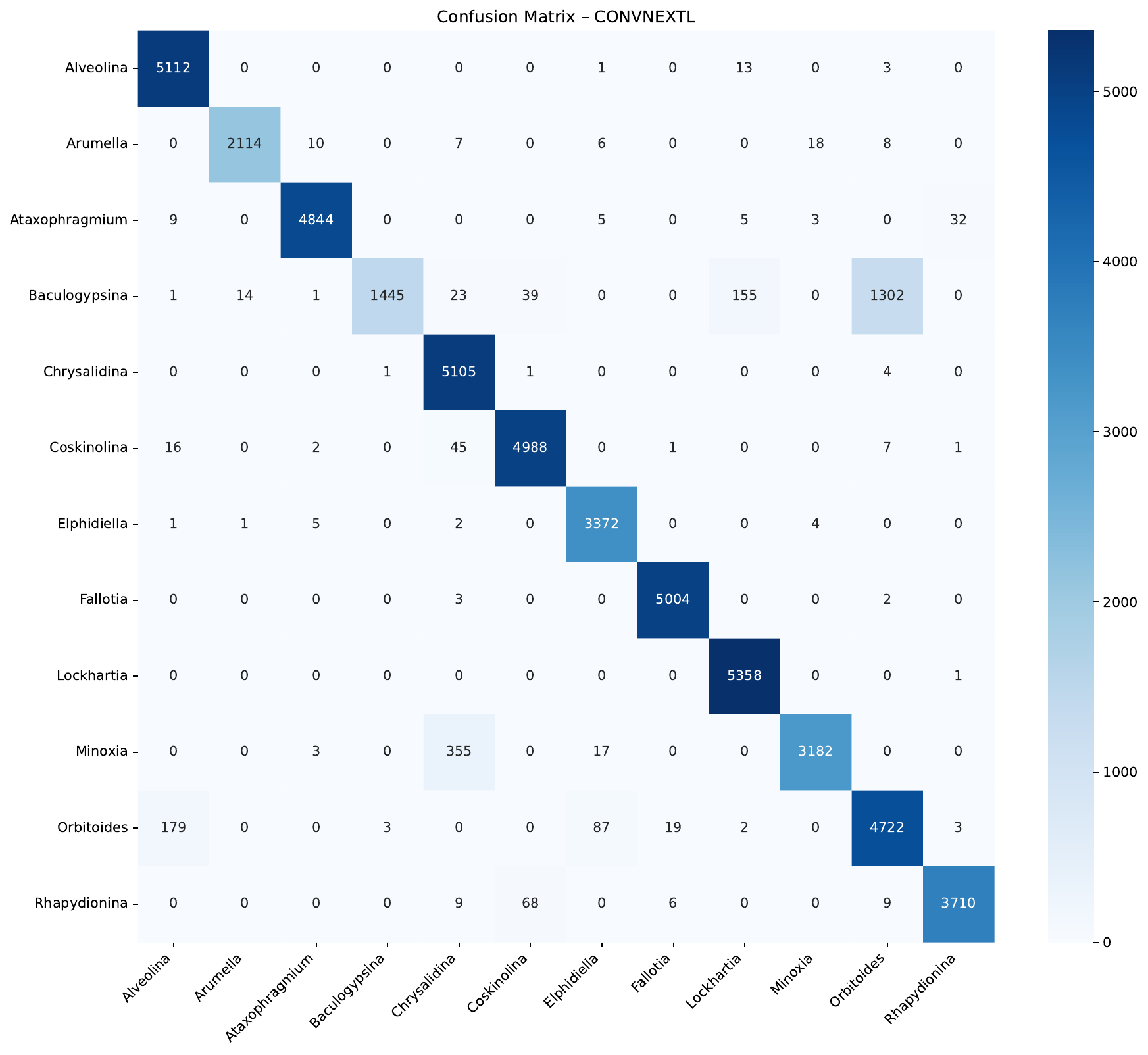}
    \subcaption{ConvNeXt-Large (best individual model).}
    \label{fig:cm_convnextl}
\end{minipage}%
\hfill
\begin{minipage}[t]{0.48\textwidth}
    \centering
    \includegraphics[width=\linewidth]{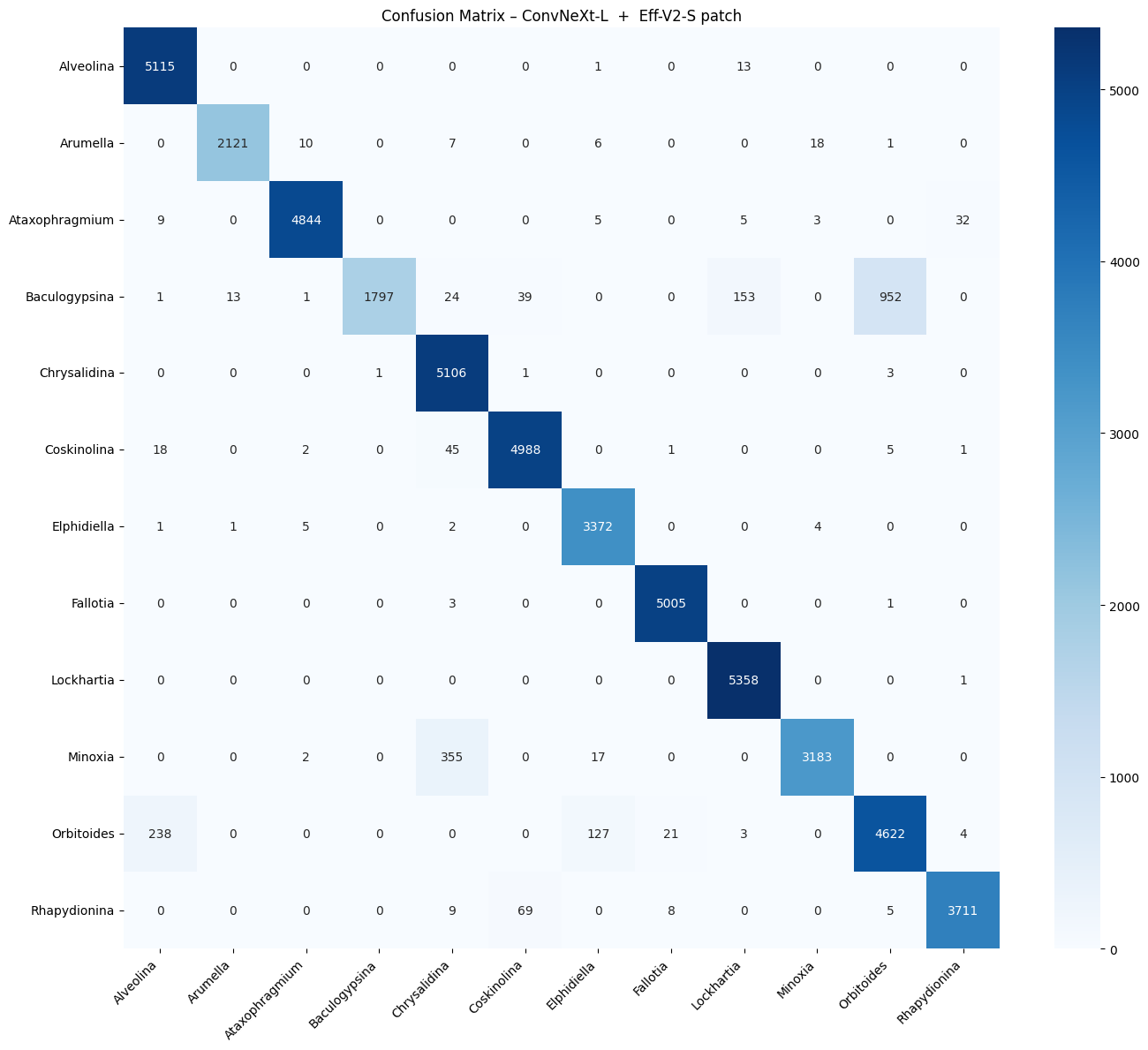}
    \subcaption{ForamDeepSlice (FDS, PatchEnsemble).}
    \label{fig:cm_fds}
\end{minipage}
\caption{
\textbf{Confusion matrices comparing the best individual model with ForamDeepSlice (FDS).}
The most prominent misclassification pattern is the confusion of true \textit{Baculogypsina} specimens as \textit{Orbitoides} (Baculogypsina row, Orbitoides column). Under ConvNeXt-Large alone~(a), 1{,}302 of 2{,}980 \textit{Baculogypsina} slices were assigned to \textit{Orbitoides}. Note the reduction in \textit{Baculogypsina} misclassification from 1{,}302 to 952 under the FDS PatchEnsemble~(b), while correct \textit{Baculogypsina} predictions simultaneously increased from 1{,}445 to 1{,}797. All other species maintain near-perfect diagonal dominance in both models.
}
\label{fig:confusion_matrices}
\end{figure}

A closer inspection of precision and recall revealed distinct error patterns (Tables~\ref{tab:precision_all} and~\ref{tab:recall_all}). \textit{Baculogypsina} consistently showed the lowest recall across models, indicating a high rate of false negatives: the model frequently missed true instances of this species. In contrast, \textit{Orbitoides} suffered from low precision, suggesting that other species were frequently misclassified as \textit{Orbitoides}, leading to false positives. Inspection of the confusion matrices (Figure~\ref{fig:confusion_matrices}a) reveals that the dominant misclassification pathway is the confusion of true \textit{Baculogypsina} specimens as \textit{Orbitoides}: under ConvNeXt-Large alone, 1{,}302 of 2{,}980 \textit{Baculogypsina} slices (43.7\%) were erroneously assigned to \textit{Orbitoides}, which simultaneously inflated \textit{Orbitoides} false positives and depressed \textit{Baculogypsina} recall. These deficiencies are reflected in their reduced F1-scores (Table~\ref{tab:f1_scores_all}), which signal a decline in overall classification quality for these species.

\begin{table}[!htbp]
\centering
\caption{Comprehensive precision values for all species across individual models and ensemble methods. Best value per species is highlighted. FDS = ForamDeepSlice (Patched Ensemble).}
\label{tab:precision_all}
\vspace{0.5em}
\footnotesize
\setlength{\tabcolsep}{3pt}
\renewcommand{\arraystretch}{1.3}
\begin{tabular}{@{}l ccccccc cccc@{}}
\toprule
& \multicolumn{7}{c}{\textbf{Individual Models}} & \multicolumn{4}{c}{\textbf{Ensemble Models}} \\
\cmidrule(lr){2-8} \cmidrule(l){9-12}
\textbf{Species} & \rotatebox{70}{ConvNeXt-L} & \rotatebox{70}{NASNet} & \rotatebox{70}{EffNetV2-L} & \rotatebox{70}{EffNetV2-S} & \rotatebox{70}{ConvNeXt} & \rotatebox{70}{MobileNet} & \rotatebox{70}{ResNet101V2} & \rotatebox{70}{Top-2} & \rotatebox{70}{Top-3} & \rotatebox{70}{Top-4} & \rotatebox{70}{FDS} \\
\midrule
Alveolina        & 0.961 & 0.934 & 0.929 & 0.944 & 0.951 & 0.969 & \best{0.984} & 0.946 & 0.952 & 0.938 & 0.950 \\
Arumella         & 0.993 & \best{0.998} & 0.996 & 0.996 & 0.996 & 0.956 & 0.985 & 0.997 & \best{0.998} & \best{0.998} & 0.993 \\
Ataxophragmium   & 0.995 & 0.992 & 0.995 & 0.991 & 0.992 & 0.930 & 0.831 & 0.992 & 0.995 & 0.995 & \best{0.995} \\
Baculogypsina    & 0.997 & 0.974 & 0.919 & 0.996 & 0.995 & 0.986 & 0.960 & 0.971 & 0.957 & 0.985 & \best{0.999} \\
Chrysalidina     & 0.920 & 0.910 & 0.945 & 0.922 & 0.942 & 0.908 & 0.612 & 0.914 & \best{0.951} & 0.950 & 0.919 \\
Coskinolina      & 0.978 & 0.954 & 0.971 & 0.980 & \best{0.996} & 0.886 & 0.962 & 0.975 & 0.975 & 0.981 & 0.978 \\
Elphidiella      & 0.966 & 0.931 & \best{0.973} & 0.773 & 0.850 & 0.623 & 0.771 & 0.910 & 0.942 & 0.949 & 0.955 \\
Fallotia         & 0.994 & 0.998 & 0.970 & 0.994 & 0.996 & 0.999 & \best{0.999} & 0.997 & 0.986 & 0.993 & 0.994 \\
Lockhartia       & 0.968 & 0.978 & 0.992 & 0.997 & 0.964 & \best{0.999} & 0.975 & 0.969 & 0.990 & 0.995 & 0.968 \\
Minoxia          & \best{0.992} & 0.968 & 0.986 & 0.808 & 0.889 & 0.885 & 0.767 & 0.973 & 0.987 & 0.988 & \best{0.992} \\
Orbitoides       & 0.779 & 0.739 & 0.690 & 0.716 & 0.626 & 0.600 & 0.603 & 0.684 & 0.666 & 0.722 & \best{0.827} \\
Rhapydionina     & 0.990 & 0.994 & 0.986 & 0.991 & 0.984 & 0.990 & 0.982 & \best{0.995} & \best{0.995} & 0.991 & 0.989 \\
\midrule
\textbf{Macro Prec.} & 0.961 & 0.948 & 0.946 & 0.926 & 0.932 & 0.894 & 0.869 & 0.944 & 0.950 & 0.957 & \best{0.963} \\
\bottomrule
\end{tabular}
\end{table}

\begin{table}[!htbp]
\centering
\caption{Comprehensive recall values for all species across individual models and ensemble methods. Best value per species is highlighted. FDS = ForamDeepSlice (Patched Ensemble).}
\label{tab:recall_all}
\vspace{0.5em}
\footnotesize
\setlength{\tabcolsep}{3pt}
\renewcommand{\arraystretch}{1.3}
\begin{tabular}{@{}l ccccccc cccc@{}}
\toprule
& \multicolumn{7}{c}{\textbf{Individual Models}} & \multicolumn{4}{c}{\textbf{Ensemble Models}} \\
\cmidrule(lr){2-8} \cmidrule(l){9-12}
\textbf{Species} & \rotatebox{70}{ConvNeXt-L} & \rotatebox{70}{NASNet} & \rotatebox{70}{EffNetV2-L} & \rotatebox{70}{EffNetV2-S} & \rotatebox{70}{ConvNeXt} & \rotatebox{70}{MobileNet} & \rotatebox{70}{ResNet101V2} & \rotatebox{70}{Top-2} & \rotatebox{70}{Top-3} & \rotatebox{70}{Top-4} & \rotatebox{70}{FDS} \\
\midrule
Alveolina        & 0.996 & 0.991 & 0.996 & 0.999 & 0.997 & \best{0.999} & 0.961 & 0.994 & 0.997 & \best{0.999} & 0.997 \\
Arumella         & 0.977 & 0.943 & \best{0.980} & 0.855 & 0.934 & 0.956 & 0.880 & 0.941 & 0.967 & 0.975 & \best{0.980} \\
Ataxophragmium   & 0.989 & 0.985 & 0.974 & 0.972 & \best{0.994} & 0.987 & 0.974 & 0.989 & 0.993 & 0.992 & 0.989 \\
Baculogypsina    & 0.484 & 0.441 & 0.330 & 0.497 & 0.080 & 0.288 & 0.105 & 0.243 & 0.226 & 0.419 & \best{0.603} \\
Chrysalidina     & 0.998 & 0.992 & 0.997 & 0.910 & 0.932 & 0.853 & 0.762 & 0.993 & 0.996 & 0.998 & \best{0.999} \\
Coskinolina      & \best{0.985} & 0.932 & 0.956 & 0.938 & 0.932 & 0.973 & 0.928 & 0.926 & 0.950 & 0.949 & \best{0.985} \\
Elphidiella      & 0.996 & 0.994 & 0.997 & 0.990 & 0.996 & 0.995 & 0.965 & 0.995 & 0.996 & \best{0.997} & 0.996 \\
Fallotia         & 0.999 & 0.999 & 0.997 & 0.999 & 0.999 & 0.998 & 0.994 & \best{0.999} & 0.999 & 0.999 & 0.999 \\
Lockhartia       & \best{0.999} & 0.999 & 0.999 & 0.998 & 0.999 & 0.992 & 0.996 & 0.999 & \best{0.999} & \best{0.999} & \best{0.999} \\
Minoxia          & 0.894 & 0.959 & \best{0.990} & 0.959 & 0.982 & 0.900 & 0.491 & 0.969 & \best{0.990} & 0.982 & 0.894 \\
Orbitoides       & \best{0.941} & 0.897 & 0.898 & 0.762 & 0.849 & 0.590 & 0.791 & 0.900 & 0.912 & 0.904 & 0.921 \\
Rhapydionina     & 0.975 & 0.926 & 0.907 & 0.970 & \best{0.997} & 0.859 & 0.944 & 0.960 & 0.949 & 0.963 & 0.976 \\
\midrule
\textbf{Macro Rec.} & 0.936 & 0.921 & 0.919 & 0.904 & 0.891 & 0.866 & 0.816 & 0.909 & 0.915 & 0.931 & \best{0.945} \\
\bottomrule
\end{tabular}
\end{table}

\subsubsection{Morphological Sources of Misclassification}

\begin{figure}[!htbp]
\centering
\begin{minipage}[c]{0.48\textwidth}
    \centering
    \includegraphics[width=\linewidth]{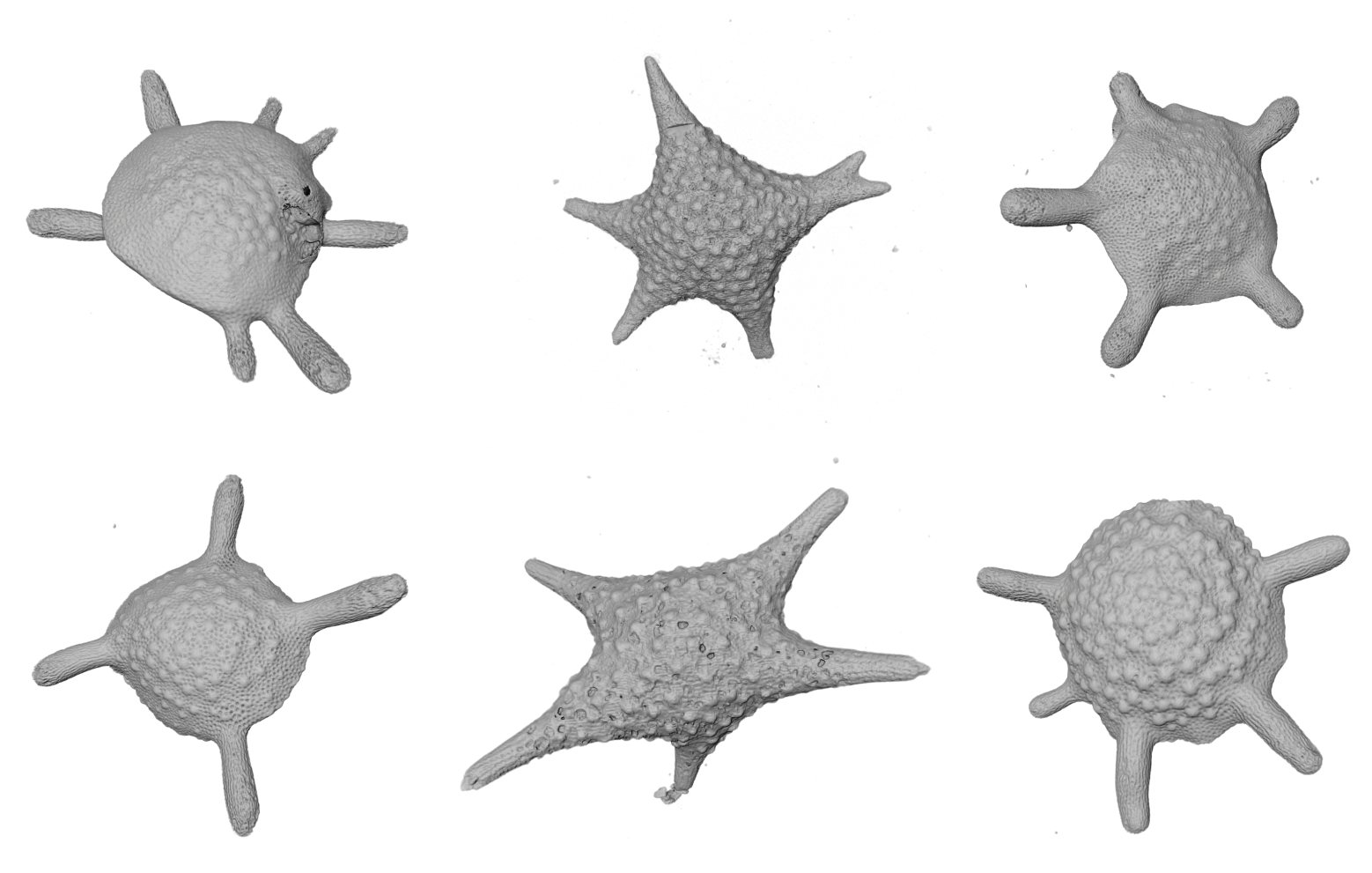}
    \subcaption{ Feature variability in \textit{Baculogypsina}.}

    Random slice sampling captures highly variable features due to the species' asymmetry, such as differing numbers of radial extensions, complicating deep learning classification.
    \label{fig:bacu}
\end{minipage}%
\hfill
\begin{minipage}[c]{0.48\textwidth}
    \centering
    \includegraphics[width=\linewidth]{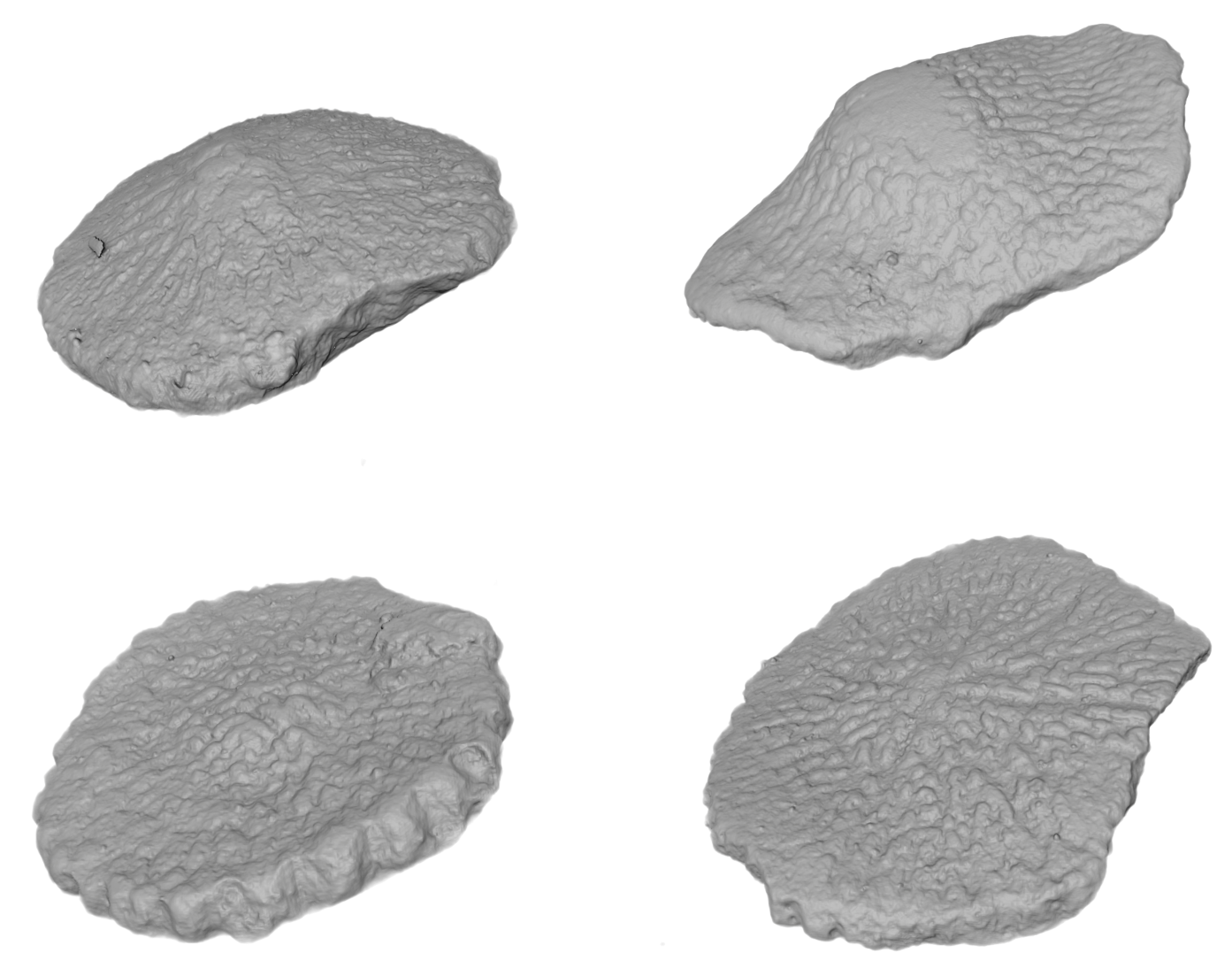}
    \subcaption{ Extraction-induced damage in \textit{Orbitoides}.}
    Physical deformations negatively impact classification accuracy. Deformations we found in forams include fractures, missing fragments, and cracks.
    \label{fig:orbi}
\end{minipage}

\caption{
\textbf{Challenges in micro-CT slice-based classification of foraminifera.}
(a) Intrinsic morphological variability in \textit{Baculogypsina} introduces inconsistencies across slices.
(b) Specimen damage in \textit{Orbitoides} further complicates classification, highlighting the need for robust strategies such as ForamDeepSlice (FDS).
}
\label{fig:classification-challenges}
\end{figure}

The classification difficulties for \textit{Baculogypsina} and \textit{Orbitoides} have clear morphological origins. Unlike the more spherical or lenticular taxa in this dataset, \textit{Baculogypsina} possesses a strongly asymmetric test with prominent radial extensions (spines) that vary in number and length among specimens. When sliced in 2D, this geometry produces highly variable cross-sections, ranging from compact circular profiles to elongated multi-armed shapes, resulting in large intra-class variance that renders the species morphologically ambiguous to the classifier (Figure~\ref{fig:classification-challenges}a). The reciprocal confusion between \textit{Baculogypsina} and \textit{Orbitoides} is further explained by their shared lenticular gross morphology, which produces similar-looking cross-sections in certain orientations. Additionally, specimen-level damage in \textit{Orbitoides}, including fractures, missing fragments, and cracks introduced during extraction (Figure~\ref{fig:classification-challenges}b), further degrades classification accuracy.

\begin{figure}[!htbp]
\centering
\includegraphics[width=\linewidth]{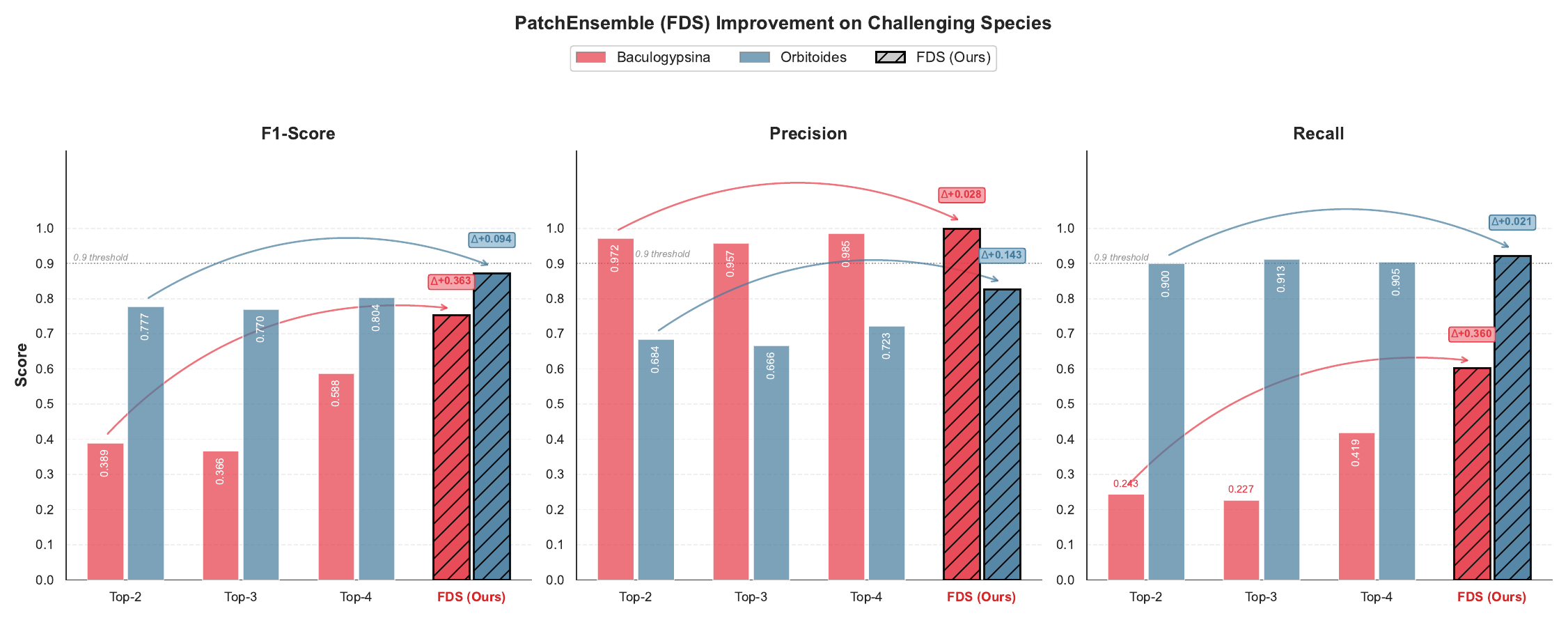}
\caption{\textbf{ForamDeepSlice (FDS) improves classification of difficult species.}
    Compared to conventional ensemble methods, the FDS PatchEnsemble strategy significantly boosts precision and recall for \textit{Baculogypsina} and \textit{Orbitoides}, which are otherwise prone to misclassification.}
\label{fig:metrics_two_species}
\end{figure}

These morphological ambiguities are reflected in a characteristic precision--recall asymmetry across all model and ensemble configurations: \textit{Baculogypsina} exhibited exceptionally high precision (up to 0.999; Table~\ref{tab:precision_all}), yet suffered from markedly low recall (as low as 0.227 under Top-3; Table~\ref{tab:recall_all}), indicating that models were highly conservative in predicting this class and frequently missed true instances. Conversely, \textit{Orbitoides} maintained strong recall values (above 0.90; Table~\ref{tab:recall_all}), but its precision was comparatively low (e.g., 0.666 in Top-3; Table~\ref{tab:precision_all}), confirming a systematic tendency to absorb false positives from misclassified \textit{Baculogypsina} specimens (Figure~\ref{fig:metrics_two_species}). These biological sources of difficulty motivate the targeted PatchEnsemble correction strategy described in the following section.

\subsection{ForamDeepSlice Ensemble Performance}

To address the systematic misclassification of \textit{Baculogypsina} and \textit{Orbitoides} identified above, we implemented a targeted ensemble strategy. Rather than relying on conventional majority voting or model averaging, we employed a \textit{PatchEnsemble} approach based on conditional, confidence-gated model switching (see Algorithm~\ref{alg:patchensemble} in Methods for the formal definition). Using the F1-score as a diagnostic metric, we identified the species with the highest misclassification rates, namely \textit{Baculogypsina} and \textit{Orbitoides}, and designated them as ``weak classes'' ($\mathcal{W}$). At inference time, the primary model (ConvNeXt-Large) generates a prediction; if that prediction falls within $\mathcal{W}$ \textit{and} the secondary patch model (EfficientNetV2-Small) exhibits higher confidence, the patch model's output replaces the primary prediction. EfficientNetV2-Small was selected as the patch model because it achieved the highest recall for \textit{Baculogypsina} (0.497; Table~\ref{tab:recall_all}) among all seven architectures evaluated. This selective switching mechanism ``patches'' the weaknesses of the primary model without degrading its strong performance on other taxa.

We compared the performance of traditional Top-$k$ ensemble combinations, Top-2 (ConvNeXt-Large + NASNet), Top-3 (Top-2 + EfficientNetV2-Large), and Top-4 (Top-3 + EfficientNetV2-Small), against our PatchEnsemble model, \textit{ForamDeepSlice} (FDS), which combines ConvNeXt-Large and EfficientNetV2-Small via confidence-gated switching. This configuration achieved 95.64\% test accuracy and 99.6\% top-3 accuracy (Table~\ref{tab:model_performance_all}). Notably, conventional Top-$k$ ensembles exacerbated the \textit{Baculogypsina}$\rightarrow$\textit{Orbitoides} confusion identified in the species-level analysis (e.g., Top-2: 2{,}041 misclassifications, 68.5\% of the class), because the additional models reinforced the same error mode. By contrast, the PatchEnsemble strategy mitigated these issues through its conditional model-switching mechanism (Algorithm~\ref{alg:patchensemble}): under FDS, \textit{Baculogypsina}$\rightarrow$\textit{Orbitoides} misclassifications dropped to 952 (31.9\%), and correct \textit{Baculogypsina} predictions rose from 1{,}445 to 1{,}797 (Figure~\ref{fig:confusion_matrices}b). This led to improved recall for \textit{Baculogypsina} (0.603 vs.\ 0.243 in Top-2; Table~\ref{tab:recall_all}) and better precision for \textit{Orbitoides} (0.827 vs.\ 0.684 in Top-2; Table~\ref{tab:precision_all}), yielding the best F1-scores of 75.2\% and 87.2\% for these two difficult species, respectively (Table~\ref{tab:baculo_orbitoides_metrics}). This targeted enhancement underscores the advantage of confidence-gated model switching over indiscriminate model aggregation in fine-grained biological classification tasks.

\begin{table}[!htbp]
\centering
\caption{Performance metrics for \textit{Baculogypsina} and \textit{Orbitoides} across ensemble methods.}
\label{tab:baculo_orbitoides_metrics}
\vspace{0.5em}
\renewcommand{\arraystretch}{1.4}
\setlength{\tabcolsep}{12pt}
\begin{tabular}{@{}lccc@{}}
\toprule
\textbf{Ensemble Method} & \textbf{F1-score} & \textbf{Precision} & \textbf{Recall} \\
\midrule
\multicolumn{4}{c}{\textit{Baculogypsina}} \\
\midrule
FDS (Ours) & \best{0.752} & \best{0.999} & \best{0.603} \\
Top-2            & 0.389 & 0.972 & 0.243 \\
Top-3            & 0.366 & 0.957 & 0.227 \\
Top-4            & 0.588 & 0.985 & 0.419 \\
\midrule
\multicolumn{4}{c}{\textit{Orbitoides}} \\
\midrule
FDS (Ours) & \best{0.872} & \best{0.827} & \best{0.922} \\
Top-2            & 0.777 & 0.684 & 0.900 \\
Top-3            & 0.770 & 0.666 & 0.913 \\
Top-4            & 0.804 & 0.723 & 0.905 \\
\bottomrule
\end{tabular}
\end{table}

Having established the classification performance of ForamDeepSlice, we next address the practical challenge of making this pipeline accessible to domain scientists who may lack computational expertise.

\subsection{Interactive Dashboard for Expert-in-the-Loop Classification}

A persistent challenge in computational paleontology is the ``last mile'' problem: even when high-performing machine learning models exist, their adoption by domain scientists is hindered if deployment requires programming expertise or command-line interaction. Analogous efforts in adjacent fields, such as CellProfiler~\cite{carpenter_cellprofiler_2006} for cell biology image analysis and QuPath~\cite{bankhead_qupath_2017} for digital pathology, have demonstrated that interactive, code-free tools can themselves constitute significant scientific contributions by democratizing access to advanced computational methods and enabling reproducible, standardized analytical workflows. In this spirit, the dashboard presented here is not merely a software demonstration but a methodological contribution that operationalizes our classification and matching pipeline into a form directly usable by micropaleontologists without computational backgrounds.

To facilitate the practical deployment of fossil classification tools, we developed an interactive dashboard tailored to the needs of paleontologists and domain scientists. The interface is designed to be intuitive and accessible, lowering the barrier to entry for non-technical users while providing advanced functionality for expert analysis. A supplementary video demonstrates the complete dashboard and an end-to-end workflow for species classification and 3D slice matching: \url{https://youtu.be/betkJ3gsmNQ}.

The application is structured into three main layers (Figure~\ref{fig:ui}):

\begin{figure}[!htbp]
\centering
\includegraphics[width=\linewidth]{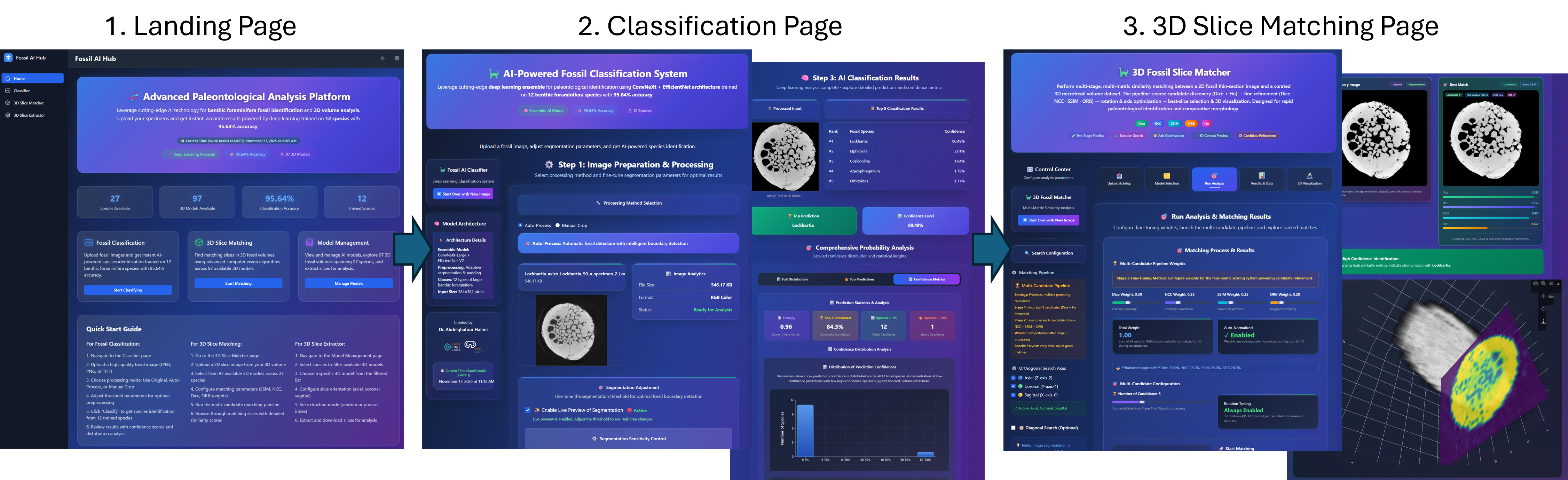}
\caption{\textbf{Our interactive dashboard provides a user-friendly access to the classifier and 3D matching algorithm.} A complete view of the dashboard in action is provided in our supplementary video (\url{https://youtu.be/betkJ3gsmNQ}).}
\label{fig:ui}
\end{figure}

\paragraph{1. Landing Page}
Upon launching the tool, users are greeted with a landing page that introduces the system's capabilities. This page includes a drag-and-drop interface for uploading micro-CT images and offers basic preprocessing options, such as contrast enhancement and image sharpening, to improve input quality prior to analysis.

\paragraph{2. Classification Page}
Once an image is uploaded and, if desired, transformed, users can navigate to the classification module. This page utilizes our FDS (PatchEnsemble-enhanced ConvNeXt-Large) model to perform species-level classification. The interface displays:
\begin{itemize}
    \item Ranked predictions with associated confidence scores.
    \item Probability distributions for each class.
    \item Visual comparisons between the input image and top-matching species.
\end{itemize}
The top-5 predictions are presented to support expert validation and further exploration, allowing users to interpret results in the context of their domain knowledge.

\paragraph{3. 3D Slice Matching Page}
Alternatively, users may choose to perform 3D slice matching. This module treats the input image as a slice from a volumetric micro-CT scan and compares it against a curated database of 3D fossil models. The matching process employs a two-stage pipeline:
\begin{itemize}
    \item \textbf{Coarse Matching:} Based on Dice coefficient and Hu moments.
    \item \textbf{Fine Matching:} Incorporates rotation-invariant analysis using SSIM\cite{zhou_wang_image_2004}, NCC\cite{Lewis_Template_1995}, and ORB feature matching\cite{rublee_orb_2011}.
\end{itemize}
Users can optionally restrict the search to a subset of models based on prior knowledge or hypotheses, enabling targeted exploration.

From a scientific perspective, the dashboard enables systematic, reproducible species identification workflows that can be applied consistently across research groups. By exposing top-$k$ predictions with calibrated confidence scores and visual comparisons against reference specimens, it supports expert-in-the-loop decision making, a paradigm increasingly recognized as essential for trustworthy AI deployment in the natural sciences. The 3D slice-matching module further enables hypothesis-driven exploration, allowing researchers to query whether an unknown specimen's internal morphology matches known reference models at specific orientations, a question directly relevant to biostratigraphic and paleoenvironmental analyses.

\subsection{Representative Workflow and Cross-Modality Case Study}

\begin{figure}[!htbp]
\centering
\includegraphics[width=\linewidth]{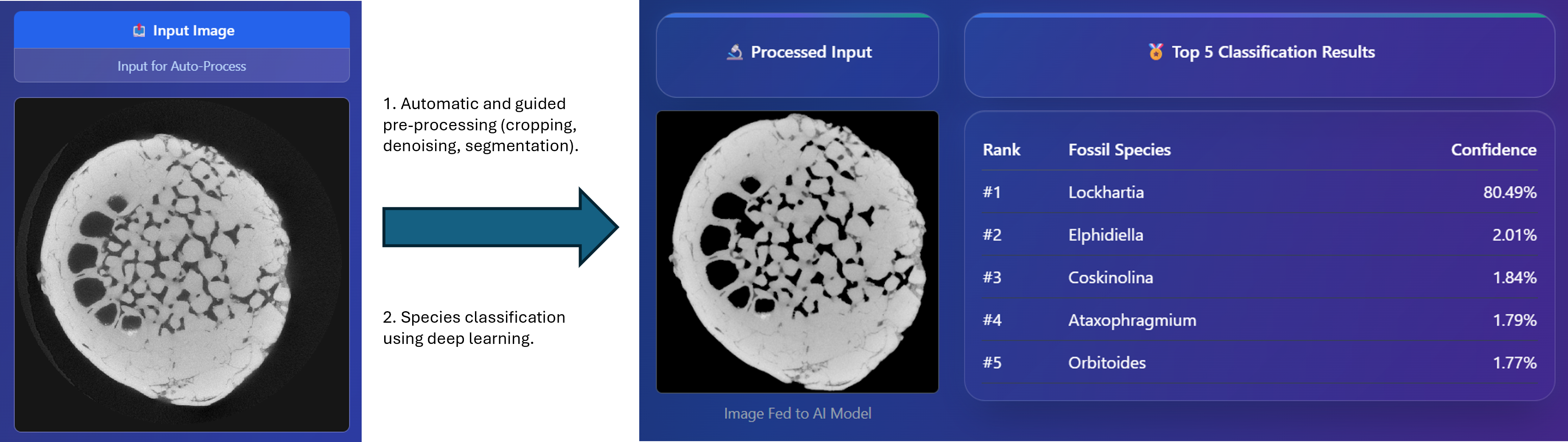}
\caption{\textbf{Classification Page.} Our AI model correctly classifies the input micro-CT image as belonging to a Lockhartia specimen.}
\label{fig:classification_step}
\end{figure}

\begin{figure}[!htbp]
\centering
\includegraphics[width=\linewidth]{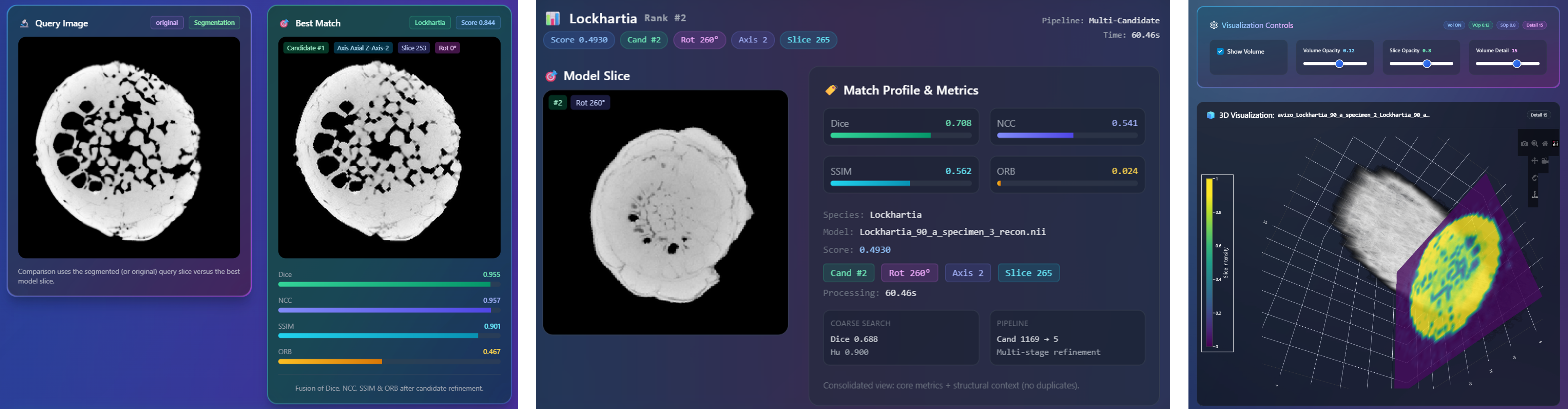}
\caption{\textbf{3D Slice Matching Page.} Our 3D matcher finds the exact slice and orientation of the input image within the sample specimen (left), but also finds the most similar slice from other samples of the same species (middle). The 3D view displays the slice in the context of the 3D sample visualization (right).}
\label{fig:matching_step}
\end{figure}

To illustrate how the dashboard and classification pipeline function in practice, we provide representative workflow examples. The first example (Figures~\ref{fig:classification_step} and~\ref{fig:matching_step}) uses a raw micro-CT slice from a \textit{Lockhartia} specimen. After dashboard-based preprocessing (cropping, denoising, and segmentation), the model correctly predicts \textit{Lockhartia} with 80.49\% confidence (Figure~\ref{fig:classification_step}). The user can then run 3D matching to recover the corresponding slice orientation within the target volume and retrieve morphologically similar slices from other specimens.

\begin{figure}[!htbp]
\centering
\includegraphics[width=\linewidth]{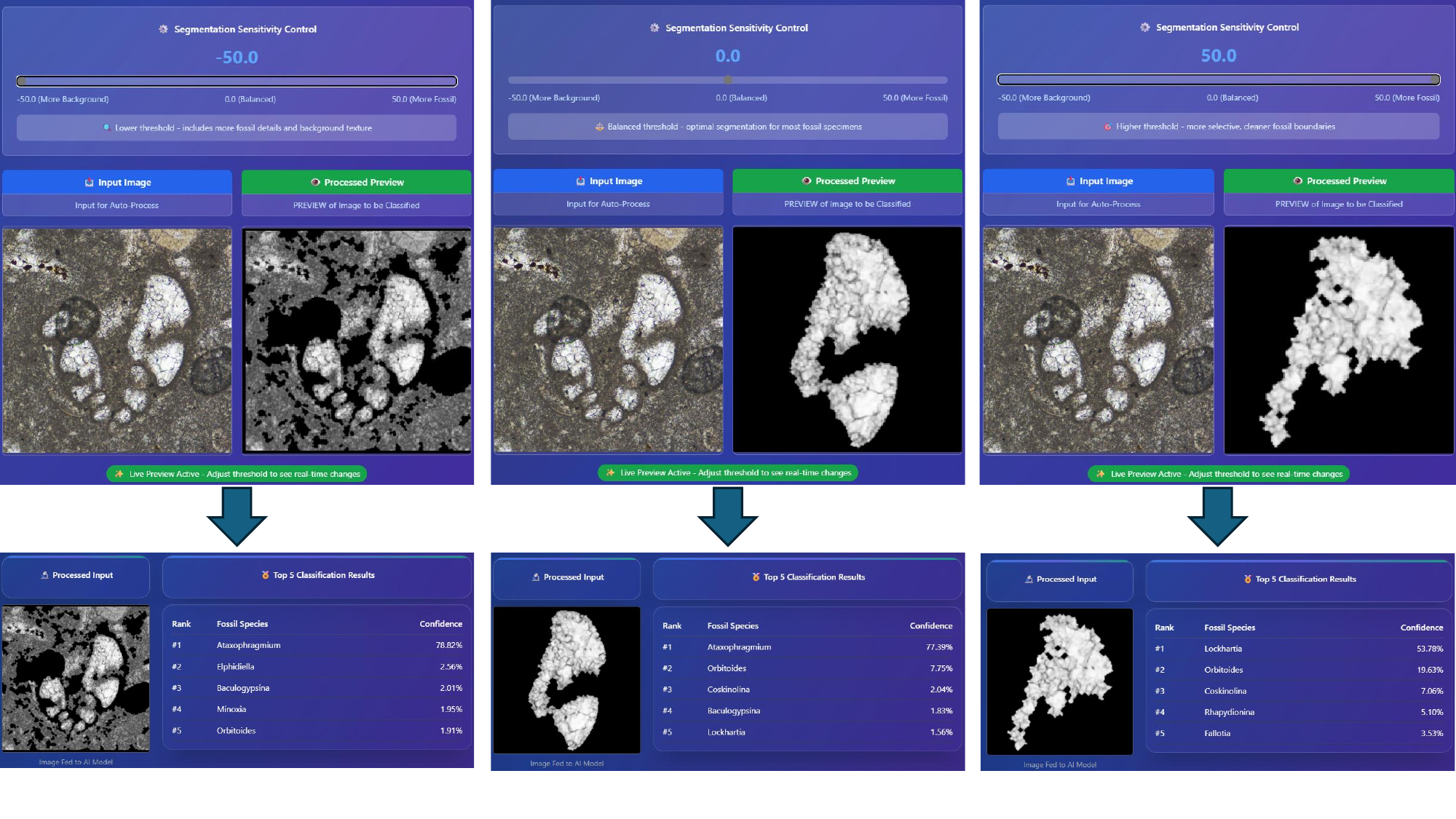}
\caption{\textbf{A qualitative optical-microscopy example showing partial cross-modality transfer of the classifier (anecdotal, not a formal benchmark).}
    This particular example also showcases how the segmentation sensitivity control affects the details retained in the input image and the final classification results. In this case, extreme background removal (rightmost) removes most of the foram features, leading to an incorrect classification, while minimal (leftmost) and balanced (middle) settings yield the correct classification of \textit{Ataxophragmium}.}
\label{fig:ataxophragmium}
\end{figure}

The second example explores cross-modality transferability. Although our framework was trained exclusively on micro-CT data, we conducted a preliminary investigation using optical microscopy images (Figure~\ref{fig:ataxophragmium}). Encouragingly, the model correctly classified the optical microscopy image of \textit{Ataxophragmium} under appropriate segmentation settings. This is most likely due to the ability of the deep learning models to learn and capture the important structures of the forams both as a whole and in parts due to their multi-scale nature. However, the introduction of artifacts such as cracks and breaks in fossils, as shown in the example, can be an interesting area to investigate and fully address in future iterations of this work. It is important to emphasize that the optical microscopy results presented here are \textit{anecdotal} and serve solely as preliminary qualitative illustrations. No systematic evaluation on a dedicated, labeled optical microscopy dataset was conducted, and the method should not be considered transferable to optical microscopy data without rigorous validation. Differences in imaging modality, including contrast mechanisms, resolution, color profiles, and the presence of preparation artifacts, may substantially affect classification performance. A formal cross-modality evaluation on a curated optical benchmark constitutes a necessary direction for future work before any claims of generalizability to non-micro-CT data can be made.

\section{Discussion}

The results presented above demonstrate that high-accuracy foraminifera classification is achievable from 2D micro-CT slices, yet several aspects of the approach and its limitations merit further discussion.
Computed tomography (CT), and especially micro-CT, has transformed paleontology by providing non-destructive access to both external and internal fossil morphology. For foraminifera, this richer structural information is particularly valuable because taxonomic discrimination often depends on internal chamber architecture that is difficult to capture reliably in conventional 2D observations.
In this study, we addressed species-level foram classification from micro-CT slices using 97 scanned specimens across 27 species, with 12 species retained for machine learning analyses based on minimum representation (at least four 3D models per species). Transfer learning with pretrained CNNs yielded strong overall performance, with ConvNeXt-Large as the best individual model. However, species-level analysis exposed persistent failure modes for \textit{Baculogypsina} and \textit{Orbitoides}. To address these cases, we introduced \textit{ForamDeepSlice} (FDS), a PatchEnsemble strategy based on conditional confidence-gated model switching between ConvNeXt-Large (main model) and EfficientNetV2-Small (patch model). Unlike majority-vote Top-$k$ ensembles, this targeted mechanism improved difficult classes without degrading performance on well-classified taxa.

To facilitate adoption by domain experts, we also developed an intuitive dashboard that enables users to classify new micro-CT images and visualize the closest matching 3D models, along with associated confidence scores. This tool is designed to lower the barrier for non-technical users and support expert-guided validation and discovery.

While this AI-based workflow represents an important step toward automated micropaleontological classification, it does not replace careful taxonomic analysis. Geological context remains indispensable, particularly in cases of homeomorphy, where unrelated foraminifera from different ages evolve similar morphologies. For example, confusion can arise between homeomorphic taxa such as \textit{Lepidorbitoides} spp. from the Cretaceous and \textit{Lepidocyclina} spp. from the Cenozoic. Although homeomorphic taxa were not included in this study, it is critical to acknowledge that the algorithm is trained only on the listed species and may misclassify specimens that resemble others outside the training set. Thus, the system should be viewed as a decision-support tool, guiding experts toward the most likely matches rather than providing definitive identifications.

An important limitation of this study concerns the generalizability of the trained models across different imaging conditions. All micro-CT data used for training, validation, and testing were acquired at a single facility (Tescan CoreTOM at KAUST Geo-Energy and Mineral Resources Platform) using consistent scanning protocols, reconstruction algorithms, and voxel resolutions. The behavior of the proposed pipeline on data acquired with different micro-CT scanners, varying tube voltages, detector configurations, voxel sizes, or reconstruction software remains untested. Domain shift arising from such variations, including differences in contrast, noise characteristics, and spatial resolution, may degrade classification accuracy. Furthermore, all specimens originate from a limited set of paleogeographic sources, and the model's robustness to specimens from geologically or geographically distinct settings has not been evaluated. Cross-device and cross-laboratory validation, potentially augmented by domain adaptation techniques, constitutes a critical direction for future work to establish the broader applicability of this framework.

In summary, this study demonstrates that combining transfer learning with a targeted confidence-gated ensemble strategy (ForamDeepSlice) can achieve high-accuracy species-level classification of foraminifera from 2D micro-CT slices. By addressing the specific failure modes of individual models through the PatchEnsemble mechanism, we obtained 95.64\% test accuracy and 99.6\% top-3 accuracy, setting a new benchmark for AI-assisted micropaleontological identification. The accompanying interactive dashboard further bridges the gap between machine learning research and routine paleontological practice by providing an accessible, code-free interface for both classification and 3D slice matching. Future work should prioritize expanding the species coverage and specimen count, conducting cross-device and cross-laboratory validation, and systematically evaluating the framework on optical microscopy data to extend its applicability to the broader micropaleontological community.

\section*{Acknowledgements}

The authors gratefully acknowledge the support of the King Abdullah University of Science and Technology (KAUST) Core Labs, particularly the Supercomputing Core Lab for providing computational resources on the Ibex cluster. We thank Domingo Lattanzi-Sanchez of KAUST Geo-Energy and Mineral Resources Platform for micro-CT scanning support. Their contributions were essential to the success of this research.

\section*{Author contributions}

A.H. conceived and conducted the experiments. R.S. and A.A. generated and curated the raw data. A.H. and D.B.B. analyzed the results and managed the project. A.M.A. supervised the project and secured funding. All authors reviewed and approved the manuscript.

\section*{Funding}
A.A. and A.M.A. acknowledge funding support from KAUST.

\section*{Competing interests}
The authors declare no competing interests.

\section*{Code and Data availability}
The test dataset is publicly available at the KAUST Research Repository~\cite{halimi_test_2026}. The full training and validation datasets, code, and trained models are available at \url{https://github.com/A-Halimi/3D_Fossil_Project}. 


\bibliographystyle{plainnat}
\bibliography{3DFossil}

@inproceedings{deng_imagenet_2009,
	address = {Miami, FL},
	title = {{ImageNet}: {A} large-scale hierarchical image database},
	isbn = {978-1-4244-3992-8},
	shorttitle = {{ImageNet}},
	url = {https://ieeexplore.ieee.org/document/5206848/},
	doi = {10.1109/CVPR.2009.5206848},
	urldate = {2025-09-08},
	booktitle = {2009 {IEEE} {Conference} on {Computer} {Vision} and {Pattern} {Recognition}},
	publisher = {IEEE},
	author = {Deng, Jia and Dong, Wei and Socher, Richard and Li, Li-Jia and {Kai Li} and {Li Fei-Fei}},
	month = jun,
	year = {2009},
	pages = {248--255},
}

@inproceedings{he_deep_2016,
	address = {Las Vegas, NV, USA},
	title = {Deep {Residual} {Learning} for {Image} {Recognition}},
	isbn = {978-1-4673-8851-1},
	url = {http://ieeexplore.ieee.org/document/7780459/},
	doi = {10.1109/CVPR.2016.90},
	urldate = {2025-09-08},
	booktitle = {2016 {IEEE} {Conference} on {Computer} {Vision} and {Pattern} {Recognition} ({CVPR})},
	publisher = {IEEE},
	author = {He, Kaiming and Zhang, Xiangyu and Ren, Shaoqing and Sun, Jian},
	month = jun,
	year = {2016},
	pages = {770--778},
}

@misc{tan_efficientnetv2_2021,
	title = {{EfficientNetV2}: {Smaller} {Models} and {Faster} {Training}},
	copyright = {arXiv.org perpetual, non-exclusive license},
	shorttitle = {{EfficientNetV2}},
	url = {https://arxiv.org/abs/2104.00298},
	doi = {10.48550/ARXIV.2104.00298},
	urldate = {2025-09-08},
	author = {Tan, Mingxing and Le, Quoc V.},
	year = {2021},
	publisher = {arXiv},
	note = {Publisher: arXiv Version Number: 3},
}

@inproceedings{liu_convnet_2022,
	address = {New Orleans, LA, USA},
	title = {A {ConvNet} for the 2020s},
	copyright = {https://doi.org/10.15223/policy-029},
	isbn = {978-1-6654-6946-3},
	url = {https://ieeexplore.ieee.org/document/9879745/},
	doi = {10.1109/CVPR52688.2022.01167},
	urldate = {2025-09-08},
	booktitle = {2022 {IEEE}/{CVF} {Conference} on {Computer} {Vision} and {Pattern} {Recognition} ({CVPR})},
	publisher = {IEEE},
	author = {Liu, Zhuang and Mao, Hanzi and Wu, Chao-Yuan and Feichtenhofer, Christoph and Darrell, Trevor and Xie, Saining},
	month = jun,
	year = {2022},
	pages = {11966--11976},
}

@inproceedings{zoph_learning_2018,
	address = {Salt Lake City, UT},
	title = {Learning {Transferable} {Architectures} for {Scalable} {Image} {Recognition}},
	isbn = {978-1-5386-6420-9},
	url = {https://ieeexplore.ieee.org/document/8579005/},
	doi = {10.1109/CVPR.2018.00907},
	urldate = {2025-09-08},
	booktitle = {2018 {IEEE}/{CVF} {Conference} on {Computer} {Vision} and {Pattern} {Recognition}},
	publisher = {IEEE},
	author = {Zoph, Barret and Vasudevan, Vijay and Shlens, Jonathon and Le, Quoc V.},
	month = jun,
	year = {2018},
	pages = {8697--8710},
}

@inproceedings{howard_searching_2019,
	address = {Seoul, Korea (South)},
	title = {Searching for {MobileNetV3}},
	copyright = {https://ieeexplore.ieee.org/Xplorehelp/downloads/license-information/IEEE.html},
	isbn = {978-1-7281-4803-8},
	url = {https://ieeexplore.ieee.org/document/9008835/},
	doi = {10.1109/ICCV.2019.00140},
	urldate = {2025-09-08},
	booktitle = {2019 {IEEE}/{CVF} {International} {Conference} on {Computer} {Vision} ({ICCV})},
	publisher = {IEEE},
	author = {Howard, Andrew and Sandler, Mark and Chen, Bo and Wang, Weijun and Chen, Liang-Chieh and Tan, Mingxing and Chu, Grace and Vasudevan, Vijay and Zhu, Yukun and Pang, Ruoming and Adam, Hartwig and Le, Quoc},
	month = oct,
	year = {2019},
	pages = {1314--1324},
}

@inproceedings{yun_cutmix_2019,
	address = {Seoul, Korea (South)},
	title = {{CutMix}: {Regularization} {Strategy} to {Train} {Strong} {Classifiers} {With} {Localizable} {Features}},
	copyright = {https://ieeexplore.ieee.org/Xplorehelp/downloads/license-information/IEEE.html},
	isbn = {978-1-7281-4803-8},
	shorttitle = {{CutMix}},
	url = {https://ieeexplore.ieee.org/document/9008296/},
	doi = {10.1109/ICCV.2019.00612},
	urldate = {2025-09-08},
	booktitle = {2019 {IEEE}/{CVF} {International} {Conference} on {Computer} {Vision} ({ICCV})},
	publisher = {IEEE},
	author = {Yun, Sangdoo and Han, Dongyoon and Chun, Sanghyuk and Oh, Seong Joon and Yoo, Youngjoon and Choe, Junsuk},
	month = oct,
	year = {2019},
	pages = {6022--6031},
}

@misc{zhang_mixup_2017,
	title = {mixup: {Beyond} {Empirical} {Risk} {Minimization}},
	copyright = {arXiv.org perpetual, non-exclusive license},
	shorttitle = {mixup},
	url = {https://arxiv.org/abs/1710.09412},
	doi = {10.48550/ARXIV.1710.09412},
	urldate = {2025-09-08},
	publisher = {arXiv},
	author = {Zhang, Hongyi and Cisse, Moustapha and Dauphin, Yann N. and Lopez-Paz, David},
	year = {2017},
	note = {Version Number: 2},
}

@article{zhou_wang_image_2004,
	title = {Image quality assessment: from error visibility to structural similarity},
	volume = {13},
	copyright = {https://ieeexplore.ieee.org/Xplorehelp/downloads/license-information/IEEE.html},
	issn = {1057-7149, 1941-0042},
	shorttitle = {Image quality assessment},
	url = {https://ieeexplore.ieee.org/document/1284395/},
	doi = {10.1109/TIP.2003.819861},
	number = {4},
	urldate = {2025-09-08},
	journal = {IEEE Transactions on Image Processing},
	author = {{Zhou Wang} and Bovik, A.C. and Sheikh, H.R. and Simoncelli, E.P.},
	month = apr,
	year = {2004},
	pages = {600--612},
}

@article{dice_measures_1945,
	title = {Measures of the {Amount} of {Ecologic} {Association} {Between} {Species}},
	volume = {26},
	copyright = {http://onlinelibrary.wiley.com/termsAndConditions\#vor},
	issn = {0012-9658, 1939-9170},
	url = {https://esajournals.onlinelibrary.wiley.com/doi/10.2307/1932409},
	doi = {10.2307/1932409},
	language = {en},
	number = {3},
	urldate = {2025-09-08},
	journal = {Ecology},
	author = {Dice, Lee R.},
	month = jul,
	year = {1945},
	pages = {297--302},
}

@article{yagis_effect_2021,
	title = {Effect of data leakage in brain {MRI} classification using {2D} convolutional neural networks},
	volume = {11},
	issn = {2045-2322},
	url = {https://www.nature.com/articles/s41598-021-01681-w},
	doi = {10.1038/s41598-021-01681-w},
	language = {en},
	number = {1},
	urldate = {2025-09-08},
	journal = {Scientific Reports},
	author = {Yagis, Ekin and Atnafu, Selamawet Workalemahu and García Seco De Herrera, Alba and Marzi, Chiara and Scheda, Riccardo and Giannelli, Marco and Tessa, Carlo and Citi, Luca and Diciotti, Stefano},
	month = nov,
	year = {2021},
	pages = {22544},
}

@inproceedings{rublee_orb_2011,
	address = {Barcelona, Spain},
	title = {{ORB}: {An} efficient alternative to {SIFT} or {SURF}},
	isbn = {978-1-4577-1102-2 978-1-4577-1101-5 978-1-4577-1100-8},
	shorttitle = {{ORB}},
	url = {http://ieeexplore.ieee.org/document/6126544/},
	doi = {10.1109/ICCV.2011.6126544},
	urldate = {2025-09-08},
	booktitle = {2011 {International} {Conference} on {Computer} {Vision}},
	publisher = {IEEE},
	author = {Rublee, Ethan and Rabaud, Vincent and Konolige, Kurt and Bradski, Gary},
	month = nov,
	year = {2011},
	pages = {2564--2571},
}

@article{otsu_threshold_1979,
	title = {A {Threshold} {Selection} {Method} from {Gray}-{Level} {Histograms}},
	volume = {9},
	issn = {0018-9472, 2168-2909},
	url = {http://ieeexplore.ieee.org/document/4310076/},
	doi = {10.1109/TSMC.1979.4310076},
	number = {1},
	urldate = {2025-09-08},
	journal = {IEEE Transactions on Systems, Man, and Cybernetics},
	author = {Otsu, Nobuyuki},
	month = jan,
	year = {1979},
	pages = {62--66},
}

@misc{loshchilov_decoupled_2017,
	title = {Decoupled {Weight} {Decay} {Regularization}},
	copyright = {arXiv.org perpetual, non-exclusive license},
	url = {https://arxiv.org/abs/1711.05101},
	doi = {10.48550/ARXIV.1711.05101},
	urldate = {2025-09-08},
	publisher = {arXiv},
	author = {Loshchilov, Ilya and Hutter, Frank},
	year = {2017},
	note = {Version Number: 3},
}

@inproceedings{szegedy_rethinking_2016,
	address = {Las Vegas, NV, USA},
	title = {Rethinking the {Inception} {Architecture} for {Computer} {Vision}},
	isbn = {978-1-4673-8851-1},
	url = {http://ieeexplore.ieee.org/document/7780677/},
	doi = {10.1109/CVPR.2016.308},
	urldate = {2025-09-08},
	booktitle = {2016 {IEEE} {Conference} on {Computer} {Vision} and {Pattern} {Recognition} ({CVPR})},
	publisher = {IEEE},
	author = {Szegedy, Christian and Vanhoucke, Vincent and Ioffe, Sergey and Shlens, Jon and Wojna, Zbigniew},
	month = jun,
	year = {2016},
	pages = {2818--2826},
}

@article{knutsen_accelerating_2024,
	title = {Accelerating segmentation of fossil {CT} scans through {Deep} {Learning}},
	volume = {14},
	issn = {2045-2322},
	url = {https://www.nature.com/articles/s41598-024-71245-1},
	doi = {10.1038/s41598-024-71245-1},
	language = {en},
	number = {1},
	urldate = {2025-09-08},
	journal = {Scientific Reports},
	author = {Knutsen, Espen M. and Konovalov, Dmitry A.},
	month = sep,
	year = {2024},
	pages = {20943},
}

@article{edie_high-throughput_2023,
	title = {High-throughput micro-{CT} scanning and deep learning segmentation workflow for analyses of shelly invertebrates and their fossils: {Examples} from marine {Bivalvia}},
	volume = {11},
	issn = {2296-701X},
	shorttitle = {High-throughput micro-{CT} scanning and deep learning segmentation workflow for analyses of shelly invertebrates and their fossils},
	url = {https://www.frontiersin.org/articles/10.3389/fevo.2023.1127756/full},
	doi = {10.3389/fevo.2023.1127756},
	urldate = {2025-09-08},
	journal = {Frontiers in Ecology and Evolution},
	author = {Edie, Stewart M. and Collins, Katie S. and Jablonski, David},
	month = mar,
	year = {2023},
	pages = {1127756},
}

@article{hou_fossil_2023,
	title = {Fossil image identification using deep learning ensembles of data augmented multiviews},
	volume = {14},
	issn = {2041-210X, 2041-210X},
	url = {https://besjournals.onlinelibrary.wiley.com/doi/10.1111/2041-210X.14229},
	doi = {10.1111/2041-210X.14229},
	language = {en},
	number = {12},
	urldate = {2025-09-08},
	journal = {Methods in Ecology and Evolution},
	author = {Hou, Chengbin and Lin, Xinyu and Huang, Hanhui and Xu, Sheng and Fan, Junxuan and Shi, Yukun and Lv, Hairong},
	month = dec,
	year = {2023},
	pages = {3020--3034},
}

@article{hou_semantic_2021,
	title = {Semantic segmentation of vertebrate microfossils from computed tomography data using a deep learning approach},
	volume = {40},
	copyright = {https://creativecommons.org/licenses/by/4.0/},
	issn = {2041-4978},
	url = {https://jm.copernicus.org/articles/40/163/2021/},
	doi = {10.5194/jm-40-163-2021},
	language = {en},
	number = {2},
	urldate = {2025-09-08},
	journal = {Journal of Micropalaeontology},
	author = {Hou, Yemao and Canul-Ku, Mario and Cui, Xindong and Hasimoto-Beltran, Rogelio and Zhu, Min},
	month = oct,
	year = {2021},
	pages = {163--173},
}

@article{itaki_automated_2020,
	title = {Automated collection of single species of microfossils using a deep learning–micromanipulator system},
	volume = {7},
	issn = {2197-4284},
	url = {https://progearthplanetsci.springeropen.com/articles/10.1186/s40645-020-00332-4},
	doi = {10.1186/s40645-020-00332-4},
	language = {en},
	number = {1},
	urldate = {2025-09-08},
	journal = {Progress in Earth and Planetary Science},
	author = {Itaki, Takuya and Taira, Yosuke and Kuwamori, Naoki and Maebayashi, Toshinori and Takeshima, Satoshi and Toya, Kenji},
	month = dec,
	year = {2020},
	pages = {19},
}

@article{itaki_innovative_2020,
	title = {Innovative microfossil (radiolarian) analysis using a system for automated image collection and {AI}-based classification of species},
	volume = {10},
	issn = {2045-2322},
	url = {https://www.nature.com/articles/s41598-020-77812-6},
	doi = {10.1038/s41598-020-77812-6},
	language = {en},
	number = {1},
	urldate = {2025-09-08},
	journal = {Scientific Reports},
	author = {Itaki, Takuya and Taira, Yosuke and Kuwamori, Naoki and Saito, Hitoshi and Ikehara, Minoru and Hoshino, Tatsuhiko},
	month = dec,
	year = {2020},
	pages = {21136},
}

@article{ozer_exploration_2024,
	title = {The exploration of the transfer learning technique for {Globotruncanita} genus against the limited low-cost light microscope images},
	volume = {18},
	issn = {1863-1703, 1863-1711},
	url = {https://link.springer.com/10.1007/s11760-024-03322-x},
	doi = {10.1007/s11760-024-03322-x},
	language = {en},
	number = {8-9},
	urldate = {2025-09-08},
	journal = {Signal, Image and Video Processing},
	author = {Ozer, Ilyas and Karaca, Ali Can and Ozer, Caner Kaya and Gorur, Kutlucan and Kocak, Ismail and Cetin, Onursal},
	month = sep,
	year = {2024},
	pages = {6363--6377},
}

@article{ozer_towards_2023,
	title = {Towards investigation of transfer learning framework for {Globotruncanita} genus and {Globotruncana} genus microfossils in {Genus}-{Level} and {Species}-{Level} prediction},
	volume = {48},
	issn = {22150986},
	url = {https://linkinghub.elsevier.com/retrieve/pii/S2215098623002677},
	doi = {10.1016/j.jestch.2023.101589},
	language = {en},
	urldate = {2025-09-08},
	journal = {Engineering Science and Technology, an International Journal},
	author = {Ozer, Ilyas and Kocak, Ismail and Cetin, Onursal and Can Karaca, Ali and Kaya Ozer, Caner and Gorur, Kutlucan},
	month = dec,
	year = {2023},
	pages = {101589},
}

@article{ozer_species-level_2023,
	title = {Species-level microfossil identification for globotruncana genus using hybrid deep learning algorithms from the scratch via a low-cost light microscope imaging},
	volume = {82},
	issn = {1380-7501, 1573-7721},
	url = {https://link.springer.com/10.1007/s11042-022-13810-2},
	doi = {10.1007/s11042-022-13810-2},
	language = {en},
	number = {9},
	urldate = {2025-09-08},
	journal = {Multimedia Tools and Applications},
	author = {Ozer, Ilyas and Ozer, Caner Kaya and Karaca, Ali Can and Gorur, Kutlucan and Kocak, Ismail and Cetin, Onursal},
	month = apr,
	year = {2023},
	pages = {13689--13718},
}

@article{gorur_species-level_2023,
	title = {Species-{Level} {Microfossil} {Prediction} for {Globotruncana} genus {Using} {Machine} {Learning} {Models}},
	volume = {48},
	issn = {2193-567X, 2191-4281},
	url = {https://link.springer.com/10.1007/s13369-022-06822-5},
	doi = {10.1007/s13369-022-06822-5},
	language = {en},
	number = {2},
	urldate = {2025-09-08},
	journal = {Arabian Journal for Science and Engineering},
	author = {Gorur, Kutlucan and Kaya Ozer, Caner and Ozer, Ilyas and Can Karaca, Ali and Cetin, Onursal and Kocak, Ismail},
	month = feb,
	year = {2023},
	pages = {1315--1332},
}

@article{hou_admorph_2020,
	title = {{ADMorph}: {A} {3D} {Digital} {Microfossil} {Morphology} {Dataset} for {Deep} {Learning}},
	volume = {8},
	copyright = {https://creativecommons.org/licenses/by/4.0/legalcode},
	issn = {2169-3536},
	shorttitle = {{ADMorph}},
	url = {https://ieeexplore.ieee.org/document/9165546/},
	doi = {10.1109/ACCESS.2020.3016267},
	urldate = {2025-09-08},
	journal = {IEEE Access},
	author = {Hou, Yemao and Cui, Xindong and Canul-Ku, Mario and Jin, Shichao and Hasimoto-Beltran, Rogelio and Guo, Qinghua and Zhu, Min},
	year = {2020},
	pages = {148744--148756},
}

@article{carvalho_automated_2020,
	title = {Automated {Microfossil} {Identification} and {Segmentation} using a {Deep} {Learning} {Approach}},
	volume = {158},
	issn = {03778398},
	url = {https://linkinghub.elsevier.com/retrieve/pii/S0377839819300830},
	doi = {10.1016/j.marmicro.2020.101890},
	language = {en},
	urldate = {2025-09-08},
	journal = {Marine Micropaleontology},
	author = {Carvalho, L.E. and Fauth, G. and Baecker Fauth, S. and Krahl, G. and Moreira, A.C. and Fernandes, C.P. and Von Wangenheim, A.},
	month = jun,
	year = {2020},
	pages = {101890},
}

@article{cunningham_virtual_2014,
	title = {A virtual world of paleontology},
	volume = {29},
	issn = {01695347},
	url = {https://linkinghub.elsevier.com/retrieve/pii/S0169534714000871},
	doi = {10.1016/j.tree.2014.04.004},
	language = {en},
	number = {6},
	urldate = {2025-09-09},
	journal = {Trends in Ecology \& Evolution},
	author = {Cunningham, John A. and Rahman, Imran A. and Lautenschlager, Stephan and Rayfield, Emily J. and Donoghue, Philip C.J.},
	month = jun,
	year = {2014},
	pages = {347--357},
}

@article{hermanova_benefits_2020,
	title = {Benefits and limits of x-ray micro-computed tomography for visualization of colonization and bioerosion of shelled organisms},
	issn = {19353952, 10948074},
	url = {https://palaeo-electronica.org/content/2020/3032-micro-ct-bioerosion},
	doi = {10.26879/1048},
	urldate = {2025-09-09},
	journal = {Palaeontologia Electronica},
	author = {Heřmanová, Zuzana and Bruthansová, Jana and Holcová, Katarína and Mikuláš, Radek and Kočová Veselská, Martina and Kočí, Tomáš and Dudák, Jan and Vohník, Martin},
	year = {2020},
}

@book{zhou2025ensemble,
  title={Ensemble methods: foundations and algorithms},
  author={Zhou, Zhi-Hua},
  year={2025},
  publisher={CRC press}
}

@inproceedings{ Lewis_Template_1995,
Author = {Lewis, JP},
Editor = {Laurendeau, D and Cedras, C},
Title = {Fast template matching},
Booktitle = {VISION INTERFACE `95, PROCEEDINGS},
Series = {PROCEEDINGS - CANADIAN IMAGE PROCESSING AND PATTERN RECOGNITION SOCIETY},
Year = {1994},
Pages = {120-123},
Note = {Vision Interface 95 Conference (VI 95), QUEBEC CITY, CANADA, MAY 16-19,
   1995},
Organization = {Canadian Image Proc \& Pattern Recognit Soc},
Publisher = {CANADIAN INFORMATION PROCESSING SOC},
Address = {430 KING ST W, STE 205, TORONTO ON M5V 1L5, CANADA},
Type = {Proceedings Paper},
Language = {English},
Affiliation = {IND LIGHT \& MAG,SAN RAFAEL,CA 94912.},
ISSN = {0843-803X},
Research-Areas = {Automation \& Control Systems; Computer Science},
Web-of-Science-Categories  = {Automation \& Control Systems; Computer Science, Artificial Intelligence},
ResearcherID-Numbers = {Lewis, J P/GYI-9327-2022},
Number-of-Cited-References = {0},
Times-Cited = {409},
Usage-Count-Last-180-days = {1},
Usage-Count-Since-2013 = {6},
Doc-Delivery-Number = {BF24A},
Web-of-Science-Index = {Conference Proceedings Citation Index - Science (CPCI-S)},
Unique-ID = {WOS:A1994BF24A00016},
DA = {2025-09-10},
}

@article{sellwood_structure_1993,
	title = {Structure and origin of limestones},
	volume = {150},
	issn = {0016-7649, 2041-479X},
	url = {https://www.lyellcollection.org/doi/10.1144/gsjgs.150.5.0801},
	doi = {10.1144/gsjgs.150.5.0801},
	language = {en},
	number = {5},
	urldate = {2025-11-16},
	journal = {Journal of the Geological Society},
	author = {Sellwood, B. W.},
	month = sep,
	year = {1993},
	pages = {801--809},
}

@article{ueno_carboniferous_2022,
	title = {Carboniferous fusuline {Foraminifera}: taxonomy, regional biostratigraphy, and palaeobiogeographic faunal development},
	volume = {512},
	issn = {0305-8719, 2041-4927},
	shorttitle = {Carboniferous fusuline {Foraminifera}},
	url = {https://www.lyellcollection.org/doi/10.1144/SP512-2021-107},
	doi = {10.1144/SP512-2021-107},
	language = {en},
	number = {1},
	urldate = {2025-11-16},
	journal = {Geological Society, London, Special Publications},
	author = {Ueno, Katsumi},
	month = jan,
	year = {2022},
	pages = {327--496},
}

@misc{halimi_test_2026,
	title = {Test Dataset for {ForamDeepSlice}},
	url = {https://repository.kaust.edu.sa/handle/10754/707843},
	doi = {10.25781/KAUST-N8900},
	publisher = {{KAUST} Research Repository},
	author = {Halimi, Abdelghafour and Alibrahim, Ali and Barradas-Bautista, Didier and Sicat, Ronell and Afifi, Abdulkader M.},
	urldate = {2026-01-05},
	date = {2026},
}

@misc{halimi_foramdeepslice_2025,
	title = {{ForamDeepSlice}: A High-Accuracy Deep Learning Framework for Foraminifera Species Classification from 2D Micro-{CT} Slices},
	rights = {Creative Commons Attribution 4.0 International},
	url = {https://arxiv.org/abs/2512.00912},
	doi = {10.48550/ARXIV.2512.00912},
	shorttitle = {{ForamDeepSlice}},
	publisher = {{arXiv}},
	author = {Halimi, Abdelghafour and Alibrahim, Ali and Barradas-Bautista, Didier and Sicat, Ronell and Afifi, Abdulkader M.},
	urldate = {2026-01-05},
	date = {2025},
	note = {Version Number: 1},
}

@ARTICLE{10171244,
  author={Ferreira-Chacua, Ivan and Koeshidayatullah, Ardiansyah Ibnu},
  journal={IEEE Access}, 
  title={ForamViT-GAN: Exploring New Paradigms in Deep Learning for Micropaleontological Image Analysis}, 
  year={2023},
  volume={11},
  number={},
  pages={67298-67307},
  doi={10.1109/ACCESS.2023.3291620}}

@book{boudagher-fadel_evolution_2018,
	title = {Evolution and {Geological} {Significance} of {Larger} {Benthic} {Foraminifera}},
	isbn = {978-1-911576-93-8},
	url = {https://www.uclpress.co.uk/products/88250},
	edition = {2nd},
	publisher = {UCL Press},
	author = {BouDagher-Fadel, Marcelle K.},
	year = {2018},
	doi = {10.14324/111.9781911576938},
	address = {London},
}

@article{zachos_trends_2001,
	title = {Trends, {Rhythms}, and {Aberrations} in {Global} {Climate} 65 {Ma} to {Present}},
	volume = {292},
	issn = {0036-8075, 1095-9203},
	url = {https://www.science.org/doi/10.1126/science.1059412},
	doi = {10.1126/science.1059412},
	number = {5517},
	journal = {Science},
	author = {Zachos, James and Pagani, Mark and Sloan, Lisa and Thomas, Ellen and Billups, Katharina},
	month = apr,
	year = {2001},
	pages = {686--693},
}

@book{jones_foraminifera_2014,
	title = {Foraminifera and their {Applications}},
	isbn = {978-1-107-03640-6},
	url = {https://www.cambridge.org/core/books/foraminifera-and-their-applications/4B1B072B1E72E10398856F88ED59D882},
	publisher = {Cambridge University Press},
	author = {Jones, Robert Wynn},
	year = {2014},
	doi = {10.1017/CBO9781139567619},
	address = {Cambridge},
}

@article{mimura_classifying_2025,
	title = {Classifying microfossil radiolarians on fractal pre-trained vision transformers},
	volume = {15},
	issn = {2045-2322},
	url = {https://www.nature.com/articles/s41598-025-90988-z},
	doi = {10.1038/s41598-025-90988-z},
	journal = {Scientific Reports},
	author = {Mimura, K. and Itaki, T. and Ikehara, M. and Hoshino, T.},
	year = {2025},
	pages = {90988},
}

@article{carpenter_cellprofiler_2006,
	title = {{CellProfiler}: image analysis software for identifying and quantifying cell phenotypes},
	volume = {7},
	issn = {1474-760X},
	url = {https://genomebiology.biomedcentral.com/articles/10.1186/gb-2006-7-10-r100},
	doi = {10.1186/gb-2006-7-10-r100},
	number = {10},
	journal = {Genome Biology},
	author = {Carpenter, Anne E. and Jones, Thouis R. and Lamprecht, Michael R. and Clarke, Colin and Kang, In Han and Friman, Ola and Guertin, David A. and Chang, Joo Han and Lindquist, Robert A. and Moffat, Jason and Golland, Polina and Sabatini, David M.},
	year = {2006},
	pages = {R100},
}

@article{bankhead_qupath_2017,
	title = {{QuPath}: {Open} source software for digital pathology image analysis},
	volume = {7},
	issn = {2045-2322},
	url = {https://www.nature.com/articles/s41598-017-17204-5},
	doi = {10.1038/s41598-017-17204-5},
	journal = {Scientific Reports},
	author = {Bankhead, Peter and Loughrey, Maurice B. and Fern\'andez, Jos\'e A. and Dombrowski, Yvonne and McArt, Darragh G. and Dunne, Philip D. and McQuaid, Stephen and Gray, Ronan T. and Murray, Liam J. and Coleman, Helen G. and James, Jacqueline A. and Salto-Tellez, Manuel and Hamilton, Peter W.},
	year = {2017},
	pages = {16878},
}

@misc{hayward_bruce_bw_world_2025,
	title = {World Foraminifera Database. Accessed at https://www.marinespecies.org/foraminifera on yyyy-mm-dd},
	rights = {Creative Commons Attribution 4.0 International},
	url = {https://www.marinespecies.org/imis.php?dasid=3113&doiid=305},
	doi = {10.14284/305},
	shorttitle = {World Foraminifera Database. Accessed at https},
	publisher = {{VLIZ}},
	author = {{Hayward, Bruce, B.W.} and {Le Coze, François, F.} and {Vachard, Daniel, D.} and {Gross, Onno, O.}},
	editora = {Consorti, Lorenzo and Hayward, Bruce W. and Le Coze, François and Vachard, Daniel and {Flanders Marine Institute (VLIZ), Belgium}},
	editoratype = {collaborator},
	urldate = {2026-02-25},
	date = {2025},
	langid = {english},
}

\end{document}